\documentclass{article}
\usepackage{arxiv}

\usepackage{geometry}
\usepackage{graphicx}
\usepackage[hidelinks]{hyperref}

\usepackage{amssymb}
\usepackage{bm}
\usepackage{lineno}
\usepackage{amsmath}

\usepackage{multirow}
\usepackage{float}
\usepackage{tabularx}
\usepackage{subcaption}
\usepackage{booktabs}
\usepackage{array,xcolor,booktabs}

\usepackage{algorithm, algpseudocode}

\usepackage{cite}
\title{CoNBONet: Conformalized Neuroscience-inspired Bayesian Operator Network for Reliability Analysis}
\author{
  Shailesh Garg  \\
  Department of Applied Mechanics\\
  Indian Institute of Technology Delhi\\
  Hauz Khas, New Delhi 110016, India. \\
  \texttt{shaileshgarg96@gmail.com} \\
  \And
  Souvik Chakraborty  \\
  Department of Applied Mechanics\\
  Yardi School of Artificial Intelligence (YScAI)\\
  Indian Institute of Technology Delhi\\
  Hauz Khas, New Delhi 110016, India. \\
  \texttt{souvik@am.iitd.ac.in}}
  
\begin{document}
\maketitle
\begin{abstract}
Time-dependent reliability analysis of nonlinear dynamical systems under stochastic excitations is a critical yet computationally demanding task. Conventional approaches, such as Monte Carlo simulation, necessitate repeated evaluations of computationally expensive numerical solvers, leading to significant computational bottlenecks. To address this challenge, we propose \textit{CoNBONet}, a neuroscience-inspired surrogate model that enables fast, energy-efficient, and uncertainty-aware reliability analysis, providing a scalable alternative to techniques such as Monte Carlo simulations. CoNBONet, short for \textbf{Co}nformalized \textbf{N}euroscience-inspired \textbf{B}ayesian
\textbf{O}perator \textbf{Net}work, leverages the expressive power of deep operator networks while integrating neuroscience-inspired neuron models to achieve fast, low-power inference. Unlike traditional surrogates such as Gaussian processes, polynomial chaos expansions, or support vector regression, that may face scalability challenges for high-dimensional, time-dependent reliability problems, CoNBONet offers \textit{fast and energy-efficient inference} enabled by a neuroscience-inspired network architecture, \textit{calibrated uncertainty quantification with theoretical guarantees} via split conformal prediction, and \textit{strong generalization capability} through an operator-learning paradigm that maps input functions to system response trajectories. Validation of the proposed CoNBONet for various nonlinear dynamical systems demonstrates that CoNBONet preserves predictive fidelity, and achieves reliable coverage of failure probabilities, making it a powerful tool for robust and scalable reliability analysis in engineering design.
\end{abstract}
\section{Introduction}\label{section: introduction}
Reliability analysis is a cornerstone of modern engineering design, enabling engineers to quantify the probability of failure of complex systems and ensure their safe and robust operation under uncertainty. Applications span diverse domains, including aerospace structures \cite{wang2025bayesian, pradlwarter2005realistic,choi2016reliability}, energy infrastructure \cite{yan2025probabilistic,parol2022reliability}, and critical civil infrastructure \cite{zhou2025assessment,benaissa2015reliability,moreira2016reliability,tabsh1991reliability}, where the ability to estimate failure probabilities directly impacts risk assessment, maintenance planning, and cost-effective decision making. For a comprehensive reliability analysis, the system's response to a variety of inputs needs to be analyzed, and with the increasing sophistication of engineering systems, characterized by high-dimensional stochastic inputs and corresponding nonlinear responses, performing rigorous reliability analysis becomes a major computational challenge. Consequently, there is a need for the development of efficient and reliable analysis frameworks.

Traditional reliability assessment methods such as the First-Order Reliability Method (FORM) \cite{maier2001first,haldar1995first,chen2025decoupled} and the Second-Order Reliability Method (SORM) \cite{hu2021second,haldar1995first,zhang2015first} have been widely applied due to their computational efficiency, but their reliance on local approximations often leads to poor accuracy for complex nonlinear dynamical systems. Monte Carlo simulations \cite{harrison2010introduction,alexander2003application,ramirez2005monte} (MCSs) remain the gold standard for reliability estimation, but its requirement for a large number of model evaluations renders it impractical for high-fidelity dynamical systems, where each evaluation involves numerically integrating a set of nonlinear ordinary or partial differential equations. This computational burden is further exacerbated for \textit{time-dependent} reliability problems, where failure probabilities must be tracked over an entire time horizon rather than at a single instant.
Surrogate modeling has emerged as a powerful strategy to alleviate the prohibitive computational cost of MCS by approximating the input–output relationship of the true system response at a fraction of the cost. Popular surrogate approaches include polynomial chaos expansions \cite{hawchar2017principal,zhang2022efficient,marelli2018active}, Gaussian process regression or Kriging \cite{kaymaz2005application,su2017gaussian,wan2025new, chakraborty2019graph}, support vector regression \cite{chen2007forecasting,yu2024dynamic,roy2019support}, polynomial correlated function approximation \cite{chakraborty2016assessment, chakraborty2016modelling, chatterjee2016bi, chakraborty2017hybrid} and response surface methods \cite{li2016response,li2023service,roussouly2013new}. While successful for problems with smooth, low-dimensional response surfaces, these methods struggle to capture the temporal dynamics and high-dimensional dependencies present in complex reliability analysis problems. Their scalability with respect to input dimensionality and their ability to generalize to unseen loading conditions remain limited.

Deep learning \cite{lecun2015deep,bengio2017deep} based surrogate modeling frameworks \cite{tang2020deep,shi2024deep,thakur2022deep,thakur2023deep} have recently shown promise in addressing these limitations by leveraging their superior approximation and generalization capabilities. In particular, operator learning architectures such as the Deep Operator Network \cite{lu2021learning,garg2022assessment} (DeepONet), Fourier Neural Operator \cite{li2020fourier} (FNO), and Wavelet Neural Operator \cite{tripura2023wavelet,garg2024neuroscience,navaneeth2025harnessing} (WNO) have demonstrated the ability to learn mappings between inputs and output function spaces, making them well-suited for dynamical systems with time-varying or stochastic excitations. These data-driven operator networks are capable of generalizing well to new input trajectories not seen during training, offering a compelling approach for time-dependent reliability problems \cite{wang2023time,navaneeth2022koopman,tripura2024multi,garg2022assessment}. However, their deployment in real-world scenarios is often constrained by the high computational and energy costs associated with conventional deep neural networks, particularly when deployed on embedded or edge computing platforms. Furthermore, in their vanilla form, the operator learning schemes fail to provide the uncertainty associated with their predictions, making their use in decision-making tasks difficult. 

To address these challenges, we propose \textbf{CoNBONet}, \textbf{Co}nformalized \textbf{N}euroscience-inspired \textbf{B}ayesian \textbf{O}perator \textbf{Net}work, a neuroscience-inspired, data-driven operator learning framework designed for energy-efficient and uncertainty-aware time-dependent reliability analysis. CoNBONet integrates the Variable Spiking Neuron (VSN) model \cite{garg2023neuroscience,garg2024neuroscience} into a DeepONet \cite{lu2021learning} architecture, leveraging the event-driven dynamics of spiking computation to significantly reduce energy consumption. In contrast to conventional continuous-activation-based neural networks, VSN-based architectures trigger computations only when spikes occur, resulting in sparse, low-latency inference that is inherently well-suited for deployment in resource-constrained environments. Furthermore, since VSNs communicate through graded spikes, their performance in regression tasks \cite{garg2023neuroscience,garg2024neuroscience,jain2025hybrid,garg2025distribution} is better than that observed when using vanilla spiking neurons. 
Now, to quantify the uncertainty associated with the predictions of this neuroscience-inspired deep operator network, the whole framework is transitioned to the Bayesian domain, resulting in the first-ever neuroscience-inspired Bayesian deep operator network. 
Finally, CoNBONet incorporates split conformal prediction \cite{angelopoulos2023conformal, lei2018distribution, shafer2008tutorial} to further calibrate the confidence intervals, enabling practitioners to quantify the uncertainty associated with the model's predictions, an essential requirement in safety-critical reliability applications. The highlights of this work are summarized as follows, 
\begin{itemize}
    \item We propose \textbf{CoNBONet}, the first neuroscience-inspired, data-driven conformalized Bayesian deep operator learning framework for time-dependent reliability analysis, combining the expressive power of deep operator networks with the energy efficiency of VSN-based computation.
    \item Through the use of split conformal prediction within the CoNBONet, we provide uncertainty bounds for the estimated probability of failure, offering a practical measure of confidence in the reliability estimates.
    \item We rigorously evaluate CoNBONet on nonlinear dynamical systems subjected to stochastic loading and demonstrate its ability to achieve and maintain predictive fidelity despite sparse event-driven communication and produce calibrated reliability estimates.
\end{itemize}

The remainder of this paper is organized as follows: Section \ref{section: problem statement} formulates the time-dependent reliability problem. Section \ref{section: proposed} presents the CoNBONet architecture and its training methodology. Section \ref{section: numerical} reports numerical experiments and compares CoNBONet against state-of-the-art surrogates. Section \ref{section: conclusion} summarizes the key findings and discusses potential directions for future work.
\section{Problem Statement}\label{section: problem statement}
Let $(\Omega,\mathcal{F},\mathbb{P})$ be a probability space and 
$\boldsymbol{\xi} : \Omega \to \mathbb{R}^d$ a random vector with distribution $\mu$. 
Consider a dynamical system subjected to stochastic loading, whose state evolution corresponding to input force $f(t;\boldsymbol{\xi})$, at time $t$ in interval $[0,T]$ is denoted by $u(t;\boldsymbol{\xi})$. 
The system is considered to have \emph{failed} whenever a user-defined performance function $m(u(t;\boldsymbol{\xi}),t)$ is negative, i.e.,
\begin{equation}
    m(u(t;\boldsymbol{\xi}),t) < 0.
\end{equation}
The performance function $m(\cdot)$ is commonly defined as the \emph{safety margin} between an allowable capacity and the actual response. For example, for a displacement threshold $u_{\mathrm{crit}}$, we may define
\begin{equation}
    m(u(t;\boldsymbol{\xi}),t) = u_{\mathrm{crit}} - u(t;\boldsymbol{\xi}),
\end{equation}
so that $m > 0$ indicates safe operation at time $t$, $m = 0$ corresponds to the onset of failure, and $m < 0$ indicates that the response has exceeded the allowable limit at the given instance $t$.
The random variable describing the \textit{first time-to-failure} (FTTF) is then given by
\begin{equation}
    \tau(\boldsymbol{\xi}) = \inf \Big\{ t \in [0,T] \;:\; m(u(t;\boldsymbol{\xi}),t) \le 0 \Big\},
\end{equation}
with the condition $\tau(\boldsymbol{\xi}) = T$, if $m(u(t;\boldsymbol{\xi}),t) > 0$ for all $t \in [0,T]$., i.e., if no failure is observed in the interval $[0,T]$. $\tau(\boldsymbol{\xi})$ represents the earliest time at which the system trajectory violates the performance criterion.
We assume that for every $\boldsymbol{\xi}$, the mapping $t \mapsto u(t;\boldsymbol{\xi})$ is continuous, ensuring that $\tau(\boldsymbol{\xi})$ is a well-defined random variable.
The \textit{time-dependent failure probability} is defined as the cumulative probability that failure has occurred by time $t$,
\begin{equation}
    P_f(t) = \mathbb{P}\big[\tau(\boldsymbol{\xi}) \le t\big].
\end{equation}
A straightforward approach to compute $P_f(t)$ is Monte Carlo simulation (MCS), which requires solving the underlying dynamical system for a large number of input realizations $\{f(t;\boldsymbol{\xi}_i)\}_{i=1}^{N}$ and then computing the first time-to-failure by monitoring the performance function over full time horizon. Mathematically this can be defined as,
\begin{equation}
    P_f(t) \simeq \widehat{P}_f^{\mathrm{MC}}(t)
    = \frac{1}{N}\sum_{i=1}^N \mathbf{1}\{\tau(\bm\xi_i) \le t\}
    = \frac{1}{N}\sum_{i=1}^N 
      \mathbf{1}\Big\{\inf_{s\in[0,t]} m\big(u(s;\boldsymbol{\xi}_i),s\big) \le 0\Big\},
\end{equation}
For nonlinear, high-dimensional, or computationally expensive models, this procedure becomes prohibitive. Moreover, in many engineering applications, such reliability estimates are required in near real-time, further motivating the need for fast and efficient surrogate models.

In this work, we seek an energy efficient,  data-driven surrogate $\mathcal{S}_\theta$ that learns the input–output operator mapping $f(t;\boldsymbol{\xi}) \mapsto u(t;\boldsymbol{\xi})$ directly from simulation or experimental data, and can predict system response corresponding to varying inputs, bypassing the need to repeatedly solve the governing equations. This surrogate is then used to construct an efficient estimator
\begin{equation}
P_f(t) \simeq\widehat{P}_f^{\mathcal{S}}(t)
= \frac{1}{N} \sum_{i=1}^{N} 
\mathbf{1}\Big\{ \inf_{s\in[0,t]} m\big(\mathcal{S}_\theta(f(\cdot;\boldsymbol{\xi}_i),s),s\big) \le 0 \Big\}.
\label{eq: surrogate probability}
\end{equation}
providing a fast and deployable approximation of $P_f(t)$. $\mathbf 1\{\cdot\}$ in Eq. \eqref{eq: surrogate probability} denotes an indicator function. Beyond accurate predictions, it is desirable that the surrogate model provide an estimate for the uncertainty associated with its predictions, so that the confidence in reliability estimates can be systematically assessed. The development of such a surrogate, combining operator learning, energy efficiency, and uncertainty quantification, is the focus of this work.
\section{Conformalized Neuroscience-Inspired Bayesian Operator Network}
\label{section: proposed}

In this section, we present \textbf{CoNBONet}, a \emph{\textbf{Co}nformalized \textbf{N}euroscience-inspired \textbf{B}ayesian \textbf{O}perator \textbf{Net}work} developed as an energy-efficient and uncertainty-aware surrogate model for time-dependent reliability analysis. The framework addresses the dual challenge of learning stochastic solution operators using sparse, event-driven computation for energy-efficient inference, while maintaining reliable uncertainty quantification.
We begin by formulating the operator learning problem of approximating a solution operator that maps stochastic input functions to corresponding system response trajectories,
\begin{equation}
    \mathcal{S}_{\boldsymbol{\theta}}:\; f(\cdot;\boldsymbol{\xi}) \mapsto u(\cdot;\boldsymbol{\xi}).
\end{equation}
Deep Operator Network \cite{lu2021learning} (DeepONet) provides a data-driven framework for learning such nonlinear operators using paired input-output samples. The DeepONet architecture consists of two subnetworks: a \emph{branch network}, which encodes the discretized input function $f(t;\boldsymbol{\xi})$ sampled at fixed sensor locations into a feature vector $\mathbf{b}$, and a \emph{trunk network}, which encodes the evaluation location $t$ into a feature vector $\mathbf{t}$. The output is obtained through their inner product,
\begin{equation}
    u(t;\boldsymbol{\xi}) \approx \mathcal{S}_{\boldsymbol{\theta}}(f,t)
    = \mathbf{b}^{\top}\mathbf{t},
    \label{eq. vanilla deeponet}
\end{equation}
where $\boldsymbol{\theta}$ denotes the trainable parameters of the branch and trunk networks. Training the operator network requires paired realizations of stochastic input functions and their corresponding system response trajectories. To generate the training dataset, given $N_s$ realizations of the stochastic input function $f^{(i)}(t;\boldsymbol{\xi}_i)$, 
we collect the corresponding system responses $u^{(i)}(t;\boldsymbol{\xi}_i)$ at 
$N_t$ discrete time points $\{t_k\}_{k=1}^{N_t}$. 
The resulting dataset can be written as
\begin{equation}
    \mathbf{D} = 
    \bigg\{
        \Big(f^{(i)}(\cdot;\boldsymbol{\xi}_i), 
        \{t_k, u^{(i)}(t_k;\boldsymbol{\xi}_i)\}_{k=1}^{N_t}\Big)
    \bigg\}_{i=1}^{N_s}.
\end{equation}
We can flatten this into a collection of triplets,
\begin{equation}
    \mathbf{D} = 
    \big\{ 
        \big(\bm{f}_n, t_n, u_n \big) 
    \big\}_{n=1}^{N_s N_t},
\end{equation}
where $\bm{f}_n$ denotes the discretized representation of the input function 
used by the branch network, $t_n$ is the corresponding time, 
and $u_n$ is the observed response.

In this work, DeepONet serves as a conceptual starting point, which is subsequently restructured to incorporate neuroscience-inspired computation and principled uncertainty modeling.
While vanilla DeepONet is effective for operator approximation, it relies on dense continuous activations, making it computationally expensive and unsuitable for deployment in energy-constrained environments.
To address this, CoNBONet replaces conventional continuous activation functions in vanilla DeepONet with {Variable Spiking Neurons} (VSNs), resulting in a neuroscience-inspired deep operator network. VSNs enable sparse, event-driven computation by transmitting information only when a learnable firing threshold is exceeded, thereby significantly reducing synaptic and memory read/ write operations and improving energy efficiency. The dynamics of the VSNs are defined as
\begin{equation}\label{eq:vsn_dyn}
\begin{gathered}
\bm M_{\bar t} = \beta \bm M_{\bar t-1} + \bm z_{\bar t}, \\
\bm y_{\bar t} =
\begin{cases}
\bm 0, & \text{if } \bm M_{\bar t} < \mathcal{T}_h, \\
\phi\left(\bm z_{\bar t}\right), & \text{if } \bm M_{\bar t} \geq \mathcal{T}_h,
\end{cases}
\end{gathered}
\end{equation}
where $\bm z_{\bar t}$ denotes the presynaptic input at spike time step $\bar t$, $\bm M_{\bar t}$ is the membrane potential, $\bm y_{\bar t}$ is the output spike, $\beta \in (0,1)$ is the membrane leakage coefficient, and $\mathcal{T}_h$ is a firing threshold that may be fixed or learned. In the neuroscience-inspired DeepONet architecture, VSNs are incorporated primarily within the branch network. Although there is no theoretical restriction on employing VSNs in the trunk network, restricting their use to the branch network provides a favorable balance between energy efficiency and predictive accuracy. In many operator-learning settings, system responses are queried for a large number of distinct input functions while the evaluation locations remain fixed; consequently, the branch network is executed repeatedly, whereas the trunk network is reused. Concentrating sparse, event-driven computation in the branch network therefore yields substantial energy savings with minimal impact on accuracy, making this architectural choice both practical and effective.

Now, the sparsity in communication induced by VSNs can potentially lead to a reduced overall accuracy of predictions; hence, it becomes necessary to quantify the uncertainty associated with the predictions of the neuroscience-inspired deep operator network. 
To achieve this, CoNBONet adopts a Bayesian formulation of the neuroscience-inspired deep operator networks, resulting in the neuroscience-inspired Bayesian operator network. Instead of learning a single fixed set of parameters, the network weights are treated as random variables drawn from a posterior distribution,
\begin{equation}
    \boldsymbol{\theta} \sim p(\boldsymbol{\theta}\mid \mathbf{D}),
\end{equation}
where $\mathbf{D}$ denotes the training dataset. In the absence of prior knowledge, a standard Gaussian prior is assumed,
\begin{equation}
    p(\boldsymbol{\theta}) = \mathcal{N}(\boldsymbol{0},\mathbb{I}).
\end{equation}
The goal of training now is to approximate the posterior distribution $p(\bm\theta|\mathbf D)$ using a variational distribution $q(\bm\theta|\bm\theta_v)$, where $\bm\theta_v$ represents the parameters of the variational distribution. In the Bayesian formulation of the neuroscience-inspired operator network, the outputs are the predicted mean $\mu_u(t,\bm f; \bm \theta)$ and standard deviation $\sigma_u(t, \bm f;\bm \theta)$ of a normal distribution corresponding to the input $f(t;\bm \xi)$ and time step $t$. This is achieved by getting two feature vectors from both the branch net and the trunk net. These are then combined using the dot product to obtain the mean and the standard deviation, respectively.
Now, the posterior distribution over parameters is given by Bayes’ rule,
\begin{equation}
    p(\boldsymbol{\theta}|\mathbf{D}) = 
    \frac{p(\mathbf{D}|\boldsymbol{\theta})\,p(\boldsymbol{\theta})}
    {\int p(\mathbf{D}|\boldsymbol{\theta})\,p(\boldsymbol{\theta})\,d\boldsymbol{\theta}},
\end{equation}
where the likelihood is factorized over independent training samples as,
\begin{equation}
    p(\mathbf{D}|\boldsymbol{\theta}) =
    \prod_{n=1}^{N_sN_t} p\bigl(u_n| t_n,\bm f_n, \boldsymbol{\theta}\bigr).
\end{equation}

In the neuroscience-inspired Bayesian operator network, the output of the network is modeled as a Gaussian distribution with predictive mean $\mu_u(t_n,\bm f_n; \bm \theta)$ and standard deviation $\sigma_u(t_n,\bm f_n; \bm \theta)$, both parameterized by the probabilistic network weights,
\begin{equation}
    p\bigl(u_n|\bm f_n,t_n,\boldsymbol{\theta}\bigr)
    = \mathcal{N}\Bigl(
    u_n \,\big|\,
    \mu_u(t_n,\bm f_n; \bm \theta),
    \,{\sigma_u}^{2}(t_n,\bm f_n; \bm \theta)\Bigr).
\end{equation}
Since the exact posterior $p(\boldsymbol{\theta}|\mathbf{D})$ is intractable, variational inference is employed to approximate it with the variational distribution $q(\boldsymbol{\theta}|\boldsymbol{\theta}_v)$, taken as a normal distribution with mean $\mu_v$ and standard deviation $\sigma_v$. The parameters $\boldsymbol{\theta}_v = (\boldsymbol{\mu}_v,\boldsymbol{\sigma}_v)$ are learned by minimizing the Kullback--Leibler (KL) divergence between $q(\boldsymbol{\theta}|\boldsymbol{\theta}_v)$ and the true posterior as,
\begin{equation}
    \mathcal{L} = \mathrm{KL}\bigl(q(\boldsymbol{\theta}|\boldsymbol{\theta}_v)
    \,\Vert\, p(\boldsymbol{\theta}|\mathbf{D})\bigr).
\end{equation}
This objective leads to evidence lower bound (ELBO) maximization, or equivalently, negative ELBO minimization. To enable backpropagation through the stochastic parameters, \textit{reparameterization trick} is employed, expressing each sample of the weights as
\begin{equation}
    \theta^{(k)} = \mu_v^{(k)} + \sigma_v^{(k)} \kappa^{(k)}, 
    \qquad \kappa^{(k)} \sim \mathcal{N}(0,1).
\end{equation}
This transforms stochastic sampling into a differentiable operation, allowing gradient-based optimization. For numerical stability, $\sigma_v^{(k)}$ is parameterized using a softplus transform,
\begin{equation}
    \sigma_v^{(k)} = \log\bigl(1 + \exp(\delta^{(k)})\bigr),
\end{equation}
ensuring strictly positive standard deviations.
For a test input $(\bm f^t,t^t)$, the predictive distribution of the neuroscience-inspired Bayesian operator is
\begin{equation}
p(u \mid \bm f^t,t^t,\mathbf D) \approx \int 
\mathcal N\big(u \mid \mu_u(t^t,\bm f^t;\bm\theta),\,\sigma_u^2(t^t,\bm f^t;\bm\theta)\big)\;q(\bm\theta)\,d\bm\theta.
\end{equation}
In practice, we approximate this integral by Monte Carlo. Draw $n_1$ posterior weight samples
$\bm\theta^{(s)}\sim q(\bm\theta)$, $s=1,\dots,n_1$, and compute
\begin{equation}
\mu^{(s)} = \mu_u(t^t,\bm f^t;\bm \theta^{(s)}),\qquad
\sigma^{(s)} = \sigma_u(t^t,\bm f^t;\bm \theta^{(s)}).
\end{equation}
Now, draw $n_2$ output samples per weight sample,
\begin{equation}
u^{(s,r)} \sim \mathcal N\big(\mu^{(s)},{\sigma^{(s)}}^2\big),\qquad r=1,\dots,n_2,
\end{equation}
and form empirical estimators,
\begin{equation}
\hat\mu=\frac{1}{n_1 n_2}\sum_{s=1}^{n_1}\sum_{r=1}^{n_2} u^{(s,r)},\qquad
\hat\sigma^2=\frac{1}{n_1 n_2}\sum_{s=1}^{n_1}\sum_{r=1}^{n_2}\big(u^{(s,r)}-\hat\mu\big)^2.
\end{equation}
The corresponding prediction interval $\mathcal{C}$, for the test input $(\bm f^t,t^t)$ is constructed as,
\begin{equation}
    \mathcal{C}(t^t, \bm f^t) = \left[\hat \mu(t^t, \bm f^t) - z \hat\sigma(t^t, \bm f^t), \, \hat\mu(t^t, \bm f^t) + z\hat\sigma(t^t, \bm f^t)\right],
\end{equation}
where $z$ is the z-score of the standard normal distribution corresponding to a given confidence level. While the neuroscience-inspired Bayesian operator network provides principled uncertainty estimates, the combined effects of variational inference, surrogate gradients, and spiking-induced sparsity prevent these estimates from guaranteeing exact finite-sample coverage. To ensure reliable uncertainty quantification with theoretical guarantees, CoNBONet, in its final form, employs split conformal prediction as a post hoc calibration step, resulting in the conformalized neuroscience-inspired Bayesian operator network.

 A calibration dataset $\mathbf D_c$, separate from the training dataset, is used for this purpose. $i$\textsuperscript{th} sample, $\{\bm f^c_i, t^c, u_i^c\}\in \mathbf D_c$, belonging to the calibration dataset is first passed through the VSN augmented Bayesian operator to obtain the mean prediction $\hat \mu(t^c, \bm f_i^c)$ and the associated standard deviation $\hat \sigma(t^c, \bm f_i^c)$. The same is repeated for the whole calibration dataset, and for each calibration sample, we compute a nonconformity score, defined as,
\begin{equation}
    e_i = \frac{\big|u^c_i - \hat\mu(t^c, \bm f_i^c)\big|}{\hat\sigma(t^c, \bm f_i^c)}.
\end{equation}
The scores $\{e_i\}$ are sorted, and a conformal parameter $q$ equal to the $(1-\alpha)$ quantile, i.e. $q = e_{\big(\lceil (n_{\mathrm{cal}}+1)(1-\alpha)\rceil/n\big)}$ is computed. $\alpha$ here is the error rate for the desired confidence interval, viz., for 95\% confidence interval, $\alpha = 0.05$. At test time, the calibrated prediction interval $\mathcal{C}_{p}$ is constructed as,
\begin{equation}
    \mathcal{C}_{p}(t^t, \bm f^t) = \left[\hat\mu(t^t, \bm f^t) - z q \hat\sigma(t^t, \bm f^t), \, \hat\mu(t^t, \bm f^t) + z q \hat\sigma(t^t, \bm f^t)\right].
\end{equation}
This calibration step theoretically guarantees that the probability that a new test output lies within the calibrated prediction interval satisfies,
\begin{equation}
    \mathbb{P}\big[u^t \in \mathcal{C}_{p}(t^t, \bm f^t)\big] \ge 1 - \alpha,
\end{equation}
under the standard exchangeability assumption of conformal prediction.
The conformal parameter $q$ obtained during calibration is specific to each time step $t$. Calibration samples are therefore selected for each time step individually, and the process is applied consistently across all time steps.
The calibrated CoNBONet surrogate is then deployed for time-dependent reliability estimation. An algorithm for training process of the CoNBONet framework is given in Algorithm \ref{alg 1}.
\begin{algorithm}[ht!]
\caption{Training CoNBONet}
\label{alg 1}
\begin{algorithmic}[1]
\Require Training dataset,  $\mathbf{D} = \big\{\big(\mathbf{f}_n, t_n, u_n \big)\big\}_{n=1}^{N_s N_t}$, calibration dataset, separately for for all time steps, $\mathbf D_c = \{\bm f^c_i, t^c, u_i^c\}_{i=1}^{N_c}$, 
hyperparameters for CoNBONet, number of epochs $N_{\text{epoch}}$.

\State Initialize trainable parameters of CoNBONet.
\For{1 to $N_{\text{epoch}}$}
       \State Forward pass through the neuroscience-inspired Bayesian operator component of CoNBONet, to compute predictive mean $\mu_u$ and standard deviation $\sigma_u$.
        \State Compute ELBO loss between variational posterior $q(\theta|\theta_v)$ and true posterior.
        \State Backpropagate using reparameterization trick and surrogate gradients for VSNs.
        \State Update parameters $\theta_v$ using suitable optimization algorithm.
\EndFor
\State \textbf{Calibration of predictive uncertainty:}
\For{each time step $t_k$}
    \State Obtain network predictions corresponding to the calibration dataset $\mathbf{D}_c$.
    \State Compute nonconformity scores $e_i$ for all calibration samples.
    \State Compute quantile $q(t_k)$ for desired confidence level $1-\alpha$.
\EndFor
\State \textbf{Output:} Trained and conformal parameter $q(t_k)$ at each time step.
\end{algorithmic}
\end{algorithm}
An algorithm for time-dependent Reliability Estimation with CoNBONet is given in Algorithm \ref{alg 2}.
\begin{algorithm}[ht!]
\caption{Time-dependent Reliability Estimation with CoNBONet}
\label{alg 2}
\begin{algorithmic}[1]
\Require Trained CoNBONet framework, set of stochastic input samples $\{f(\bm \xi_i,t)\}_{i=1}^N$, performance function $m(u(t;\bm\xi),t)$, time horizon $[0,T]$.
\State For each input sample $f(t;\xi_i)$, predict response trajectory, $\hat{u}(t,\xi_i)$  and the calibrated prediction intervals using CoNBONet.
\State Evaluate first time to failure using mean prediction, lower bound, and the upper bound of the confidence interval. 
\State Compute failure probability and the required probability density functions for the first time to failure. 
\end{algorithmic}
\end{algorithm}
\begin{figure}[ht!]
    \centering
    \includegraphics[width=1\linewidth]{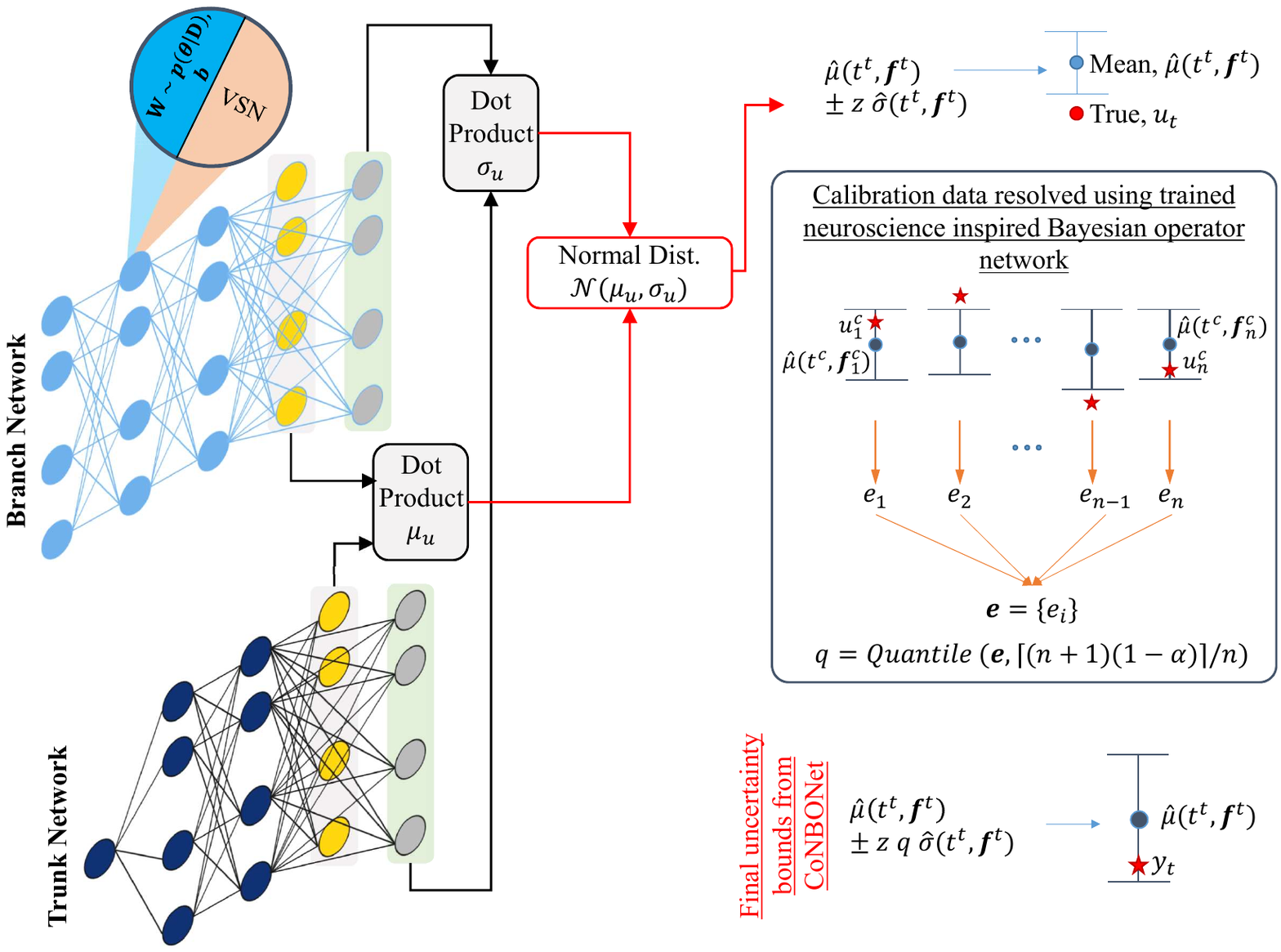}
    \caption{Schematic for information flow in CoNBONet.}
    \label{fig: flowchart}
\end{figure}
While the proposed CoNBONet framework is designed to enable energy-efficient inference through sparse, event-driven computation, it is important to analytically characterize how sparsity in communication translates into reduced computational and memory energy consumption. To this end, we next present a hardware-agnostic analytical study that quantifies the energy efficiency of Variable Spiking Neurons relative to conventional dense neural network layers.
\section{Analytical Study of Energy Efficiency of VSN Layers}\label{appendix: energy}
The primary motivation behind Variable Spiking Neurons (VSNs) is the principle of sparse communication, which is widely regarded as a key mechanism enabling energy-efficient information processing in biological neural systems \textit{in-vivo}. By restricting neural activity to discrete spike events, spiking models reduce the number of active computations, thereby offering the potential for substantially lower energy consumption.
Energy expenditure during the execution of a neural network arises from multiple sources and depends on the underlying hardware architecture, implementation strategy, and available system-level optimizations. While absolute energy measurements are therefore hardware-specific, dominant contributors to energy consumption can be analyzed in a hardware-agnostic manner. 

In this section, a principled relationship between network sparsity and energy consumption in VSN augmented networks is developed, focusing on the primary sources of computational cost, namely, arithmetic operations and memory read/write activities. This analysis highlights how sparse spiking activity directly translates into improved energy efficiency, independent of platform-specific optimizations.
The analysis compares a densely connected Artificial Neural Network (ANN) layer with a densely connected VSN layer having $N_{\text{in}}$ inputs and $N_{\text{out}}$ outputs. For the VSN layer, input activity is assumed to be sparse and distributed across $T_s$ spike time steps. The energy model adopted is intentionally conservative and does not implicitly favor either ANN or VSN execution. Unless otherwise stated, simplifying assumptions should be interpreted as worst-case scenarios.
%
Now, for both ANN and VSN layers, total energy consumption is decomposed into arithmetic computation energy and memory access energy,
\begin{equation}
E_{\text{total}} = E_{\text{ops}} + E_{\text{mem}}.
\end{equation}
This decomposition reflects modern digital hardware behavior, where multiplication operations and memory accesses dominate overall energy expenditure.

\subsection{Compute Energy: Dense versus Event-Driven Execution}
In a dense ANN layer, inference is performed using matrix--vector multiplication, where every input activation contributes to every output neuron. Consequently, the number of Multiply--Accumulate (MAC) operations is fixed and given by,
\begin{equation}
\text{MAC}_{\text{ANN}} = N_{\text{in}} \cdot N_{\text{out}}.
\end{equation}
This computation pattern does not exploit data-dependent sparsity and represents the worst-case scenario from an energy perspective.
In contrast, the VSN layer operates in an event-driven manner, where computation is triggered only by input spikes. The total number of effective input spikes over a time window $T_s$ is,
\begin{equation}
\theta_{l-1} = N_{\text{in}} \cdot T_s \cdot \alpha_{\text{in}},
\end{equation}
where $\alpha_{\text{in}}$ denotes the average input spiking activity per time step. Each incoming spike fans out to all output neurons, yielding,
\begin{equation}
\text{MAC}_{\text{VSN}}^{(\text{syn})} = \theta_{l-1} \cdot N_{\text{out}}.
\end{equation}
In addition, each output neuron incurs one multiplication per time step due to leakage dynamics,
\begin{equation}
\text{MAC}_{\text{VSN}}^{(\text{leak})} = T_s \cdot N_{\text{out}}.
\end{equation}
The total MAC count for the VSN layer is therefore
\begin{equation}
\text{MAC}_{\text{VSN}} = \text{MAC}_{\text{VSN}}^{(\text{syn})}+ \text{MAC}_{\text{VSN}}^{(\text{leak})}=
\theta_{l-1} \cdot N_{\text{out}} + T_s \cdot N_{\text{out}}.
\end{equation}
When $\alpha_{\text{in}} \ll 1$, synaptic computation in the VSN layer scales linearly with input activity rather than network size, resulting in significantly fewer multiplications than in dense ANN execution.

\subsection{Accumulation and Control Overhead}
The ANN layer incurs a small number of additional Accumulate (ACC) operations due to bias addition and indexing,
\begin{equation}
\text{ACC}_{\text{ANN}} = N_{\text{out}} + (N_{\text{in}} + N_{\text{out}}).
\end{equation}
These operations are deterministic and contribute negligibly compared to dense MAC operations.
The VSN layer requires additional accumulation operations to support neuron dynamics, including membrane integration and spike-triggered updates,
\begin{equation}
\text{ACC}_{\text{VSN}} =
2 \cdot T_s \cdot N_{\text{out}} +
\theta_{l-1} \cdot N_{\text{out}}.
\end{equation}
Although this introduces overhead absent in ANN layers, these operations scale with time or spiking activity rather than full input dimensionality. Moreover, ACC operations consume significantly less energy than MAC or memory access operations, allowing the reduction in MACs to dominate overall energy savings.

\subsection{Memory Access Energy}
Memory access energy is modeled by counting SRAM read and write operations. To establish an upper bound on VSN memory energy, all neuron state variables are conservatively assumed to be stored in SRAM and accessed at every time step.
For the ANN layer, all input activations are read once,
\begin{equation}
\text{RdIn}_{\text{ANN}} = N_{\text{in}},
\end{equation}
and all weights and biases are read once per inference,
\begin{equation}
\text{RdParam}_{\text{ANN}} = (N_{\text{in}} + 1)\cdot N_{\text{out}}.
\end{equation}
This memory access pattern is fixed and independent of input sparsity. The total read operation for ANN can thus be defined as,
\begin{equation}
    \text{Rd}_{\text{ANN}} = \text{RdIn}_{\text{ANN}} +\text{RdParam}_{\text{ANN}} =  N_{\text{in}} + (N_{\text{in}} + 1)\cdot N_{\text{out}}.
\end{equation}
In the VSN layer, inputs are read only when spikes occur,
\begin{equation}
\text{RdIn}_{\text{VSN}} = \theta_{l-1},
\end{equation}
and synaptic weights are accessed only in response to spikes,
\begin{equation}
\text{RdParam}_{\text{VSN}}^{(\text{syn})} =
(\theta_{l-1} + 1)\cdot N_{\text{out}}.
\end{equation}
This directly converts input sparsity into reduced memory access energy.
Unlike ANN pre-activations, which can be stored on local registers and discarded after use, VSN membrane potentials must persist across time steps. Conservatively, membrane potentials are assumed to be read from memory at every time step,
\begin{equation}
\text{RdParam}_{\text{VSN}}^{(\text{mem})} = T_s \cdot N_{\text{out}}.
\end{equation}
Also, the thresholds, and leakage parameters will be required to be read for a VSN layer,
\begin{equation}
\text{RdParam}_{\text{VSN}}^{(\text{param})} = 2N_{\text{out}}.
\end{equation}
The total read operation for the VSN layer can thus be defined as,
\begin{multline}
    \text{Rd}_{\text{VSN}} = \text{RdIn}_{\text{VSN}} +\text{RdParam}_{\text{VSN}}^{(\text{syn})} + \text{RdParam}_{\text{VSN}}^{(\text{mem})}+\text{RdParam}_{\text{VSN}}^{(\text{param})} =\\  \theta_{l-1} + (\theta_{l-1} + 1)\cdot N_{\text{out}}+T_s \cdot N_{\text{out}}+2N_{\text{out}}.
\end{multline}
This assumption represents a worst-case implementation; in practice, many architectures retain neuron state in registers or local buffers, substantially reducing memory energy. Therefore, this memory model is intentionally pessimistic for VSNs, thereby biasing the analysis against spiking execution and ensuring that any observed energy advantage is not an artifact of favorable assumptions.
Now, Energy is also expended on write operations. The ANN layer writes each output activation once,
\begin{equation}
\text{WrOut}_{\text{ANN}} = N_{\text{out}}.
\end{equation}
The VSN layer writes both output spikes and updated neuron states,
\begin{equation}
\text{WrOut}_{\text{VSN}} = \theta_l + T_s \cdot N_{\text{out}}.
\end{equation}
\subsection{Energy Model}

Let $E_{\text{MAC}}$ and $E_{\text{ACC}}$ denote the energy cost of a single MAC and ACC operation, respectively, and let $E_{\text{Rd}}$ and $E_{\text{Wr}}$ denote the energy cost of a single SRAM read and write. The arithmetic computation energy is modeled as
\begin{equation}
E_{\text{ops}} =
E_{\text{MAC}} \cdot \text{MAC}_i +
E_{\text{ACC}} \cdot \text{ACC}_i,\quad i\in\{\text{ANN}, \text{VSN}\}
\end{equation}
while the memory access energy is given by
\begin{equation}
E_{\text{mem}} =
E_{\text{Rd}} \cdot \text{Rd}_i +
E_{\text{Wr}} \cdot \text{WrOut}_i,\quad i\in\{\text{ANN}, \text{VSN}\}.
\end{equation}
\subsection{Numerical Analysis and Implications}
\begin{figure}[ht!]
\begin{subfigure}{0.49\textwidth}
    \centering
    \includegraphics[width = \textwidth]{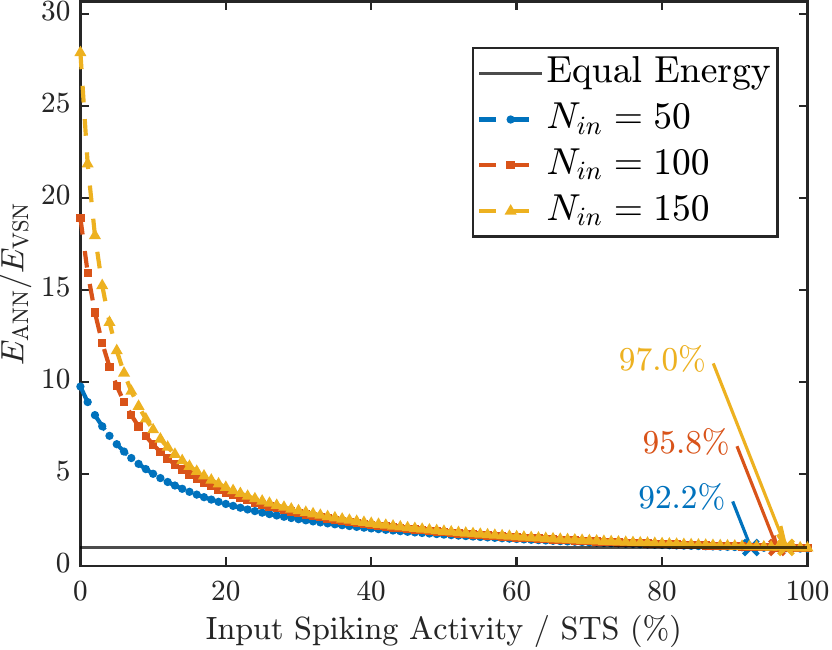}
    \caption{\centering Varying number of input nodes, Nout = 100.}
    \label{fig:energy_vs_activity_T1}
\end{subfigure}
\begin{subfigure}{0.49\textwidth}
    \centering
    \includegraphics[width = \textwidth]{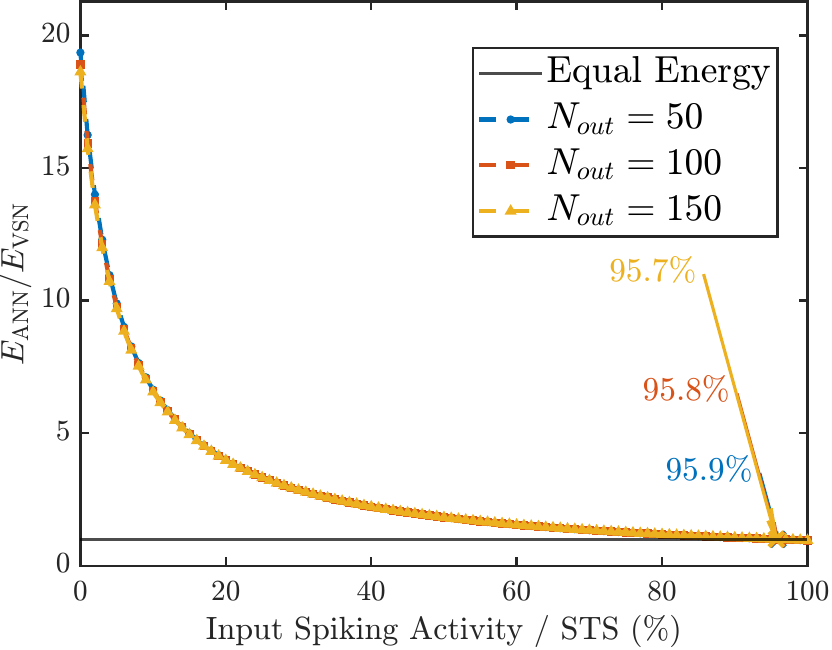}
    \caption{\centering Varying number of output nodes, Nin = 100.}
    \label{fig:energy_vs_activity_T1_out}
\end{subfigure}
\caption{Ratio of energy consumed by densely connected ANN layer vs densely connected VSN layer, when run for a single Spike Time Step (STS). It is assumed that all output nodes will produce spikes.}
\label{fig:energy_vs_activity_T1_both}
\end{figure}
\begin{figure}[ht!]
\begin{subfigure}{0.49\textwidth}
    \centering
    \includegraphics[width = \textwidth]{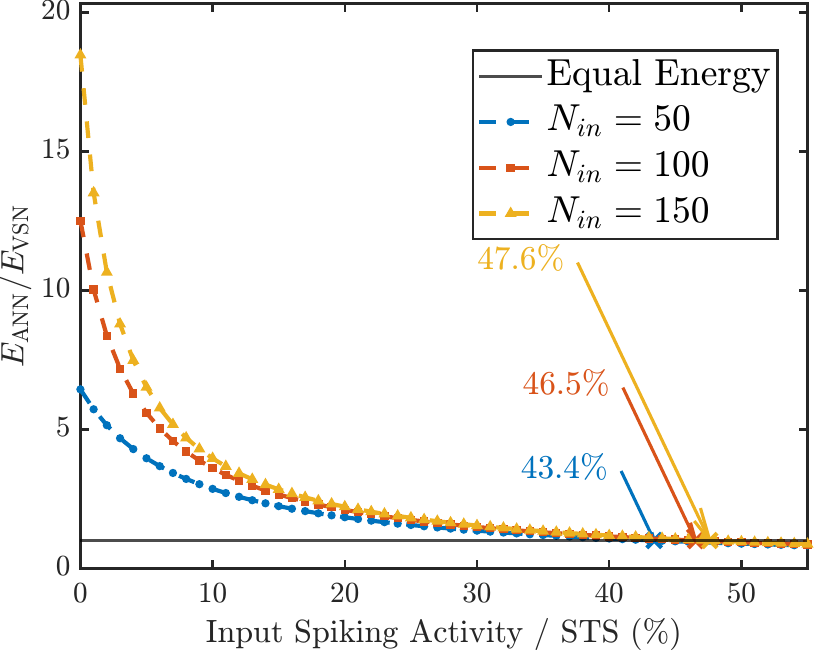}
    \caption{\centering Varying number of input nodes, Nout = 100. It is assumed that all output nodes will produce spikes.}
    \label{fig:energy_vs_activity_T2}
\end{subfigure}
\begin{subfigure}{0.49\textwidth}
    \centering
    \includegraphics[width = \textwidth]{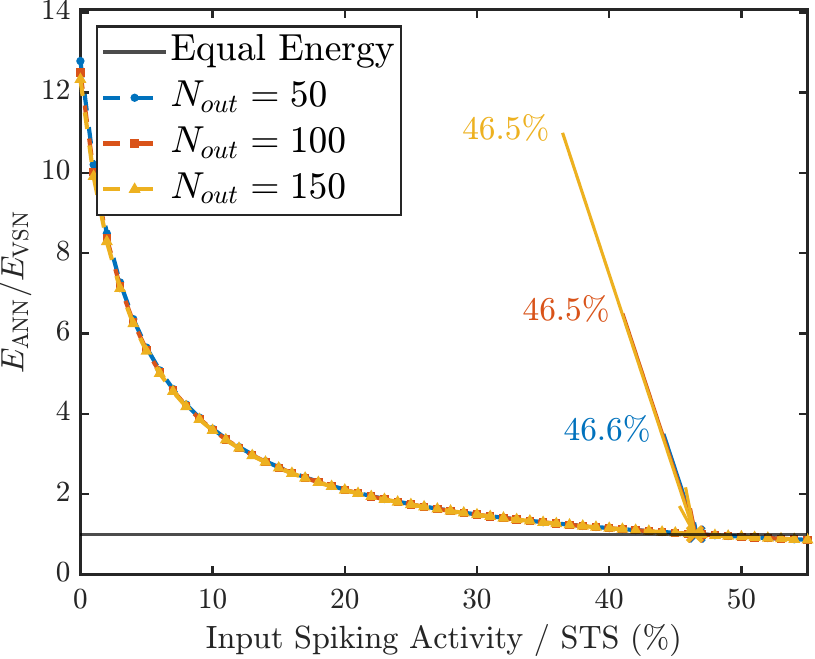}
    \caption{\centering Varying number of output nodes, Nin = 100. It is assumed that all output nodes will produce spikes.}
\label{fig:energy_vs_activity_T2_out}
\end{subfigure}
\caption{Ratio of energy consumed by densely connected ANN layer vs densely connected VSN layer, when run for two STSs.}
\label{fig:energy_vs_activity_T2_both}
\end{figure}
Figs.~\ref{fig:energy_vs_activity_T1_both} and~\ref{fig:energy_vs_activity_T2_both} illustrate the ratio of energy consumption between dense ANN and VSN layers as a function of input spiking activity for $T_s = 1$ and $T_s = 2$, respectively. Arithmetic energy parameters are set to $E_{\text{ACC}} = 0.1$~pJ and $E_{\text{MAC}} = 3.1$~pJ, while SRAM access energy is modeled as a function of memory capacity based on reported values for 45~nm CMOS technology \cite{jouppi2021ten}.

For the ANN layer, total energy consumption remains constant across activity levels. In contrast, the VSN layer exhibits strong dependence on input spiking activity. At low activity levels, synaptic MAC operations scale with $\theta_{l-1}$, resulting in substantial reductions in multiplication energy. Even after accounting for leakage-related MACs and accumulation overhead, the VSN layer remains significantly more energy-efficient for sparse inputs.
Memory access dominates energy consumption in both architectures, as a single SRAM access is one to two orders of magnitude more expensive than an addition and more than an order of magnitude more expensive than a multiplication. Although the memory model is conservatively biased against the VSN layer by assuming frequent state read/write operations, the VSN layer remains more energy-efficient than the ANN layer, provided that sufficient sparsity is achieved. Energy parity is approached only at very high spiking activity levels, $\sim$90\% for $T_s=1$ and $\sim$86\% for $T_s=2$.

Notably, this analysis does not rely on neuromorphic hardware support. On practical neuromorphic platforms, the neuron state could potentially be retained in registers or local buffers, which eliminates a dominant memory energy term and further widens the energy gap in favor of spiking computation. Additionally, activation function energy is not explicitly modeled. Since ANN activations are evaluated for all neurons whereas VSN activations are computed only upon firing, including activation energy would further favor VSN layers.

Taken together, these results demonstrate that cumulative spiking activity directly governs energy consumption. Because both arithmetic operations and memory accesses scale with the total number of spikes, total spiking activity provides a physically meaningful and practically useful proxy for energy efficiency. Accordingly, throughout this work, total spiking activity is used as a surrogate metric for energy consumption. It is worth noting that model-level optimizations such as pruning and quantization can further reduce the energy consumption of dense artificial neural networks. However, these techniques are not specific to ANN architectures and can be applied equally to spiking neural networks \cite{li2022quantization,wei2025qp,jiang2025spatio,chen2021pruning}. Since such optimizations are orthogonal to the execution paradigm, they are not explicitly considered here in order to isolate the energy implications of sparse, event-driven computation.
\section{Numerical Illustration}\label{section: numerical}
In this section, we present three examples covering Single Degree Of Freedom (DOF) and Multi DOF dynamical systems to showcase the efficacy of proposed CoNBONet for reliability analysis. The first example deals with a Single DOF system with Bouc Wen nonlinearity while the second example deals with a five DOF system with a Duffing oscillator attached to first DOF. The third example covers a 76 DOF nonlinear system, whose linear parameters are taken from a 76 DOF ASCE benchmark \cite{yang2004benchmark,nayek2019gaussian,garg2022assessment} building. The performance of the proposed CoNBONet as a surrogate framework for predicting system responses and estimating the reliability of the dynamical system is compared against that of the gold standard MCS simulations, hereafter referred to as the ground truth. Furthermore, comparisons are drawn against different variations of CoNBONet, including vanilla Bayesian Operator Network (BONet) and its conformalized counterpart Conformalized BONet (CoBONet). Comparisons are also drawn against a Neuroscience-inspired Bayesian Operator Network (NBONet), which basically utilizes VSNs within the BONet architecture. 
\subsection{Example I: single DOF base isolator system}
In this example, a single DOF Bouc-Wen system is considered, which consists of a mass mounted on a base through a nonlinear hysteretic isolator whose restoring force is defined by the Bouc-Wen model. The Bouc–Wen model has been extensively employed in the literature to characterize complex hysteretic behaviors in structural and mechanical systems, particularly in the modeling of base-isolated buildings, hysteretic dampers, and energy dissipation devices subjected to dynamic loading. The governing equation of motion for the system is defined as, 
\begin{equation}
\begin{matrix}
    m\ddot x  + c\dot x + kx + (1-k_r)Q_yz = f\\
    \dot z = \frac{1}{D_y}(\alpha\dot x - \gamma z|\dot x||z|^{\eta-1} - \beta\dot x|z|^\eta),
\end{matrix}
\end{equation}
where $m = 6800$ kg, $c = 3750$ Ns/m and $k = 2.32 \times 10^5$ N/m are the mass, damping coefficient and stiffness respectively. $Q_y = 0.05\,mg$, $k_r = 1/6$, $\alpha = 1$, $\beta=0.5$, $\gamma=0.5$, $D_y=0.0013$ and $\eta=2$ are the parameters of the Bouc-Wen oscillator. Initial conditions for data simulation are taken as $x_0 = 0.005$ m, $\dot x_0 = 0.001$ m/s and $z_0 = 0.001$. The inputs were generated using a zero mean Gaussian random field $\mathcal{GP}(\bm 0,\kappa(t,t'))$ with kernel function defined as follows,
\begin{equation}
    \kappa(t,t') = \sigma^2exp\left({-\frac{(t-t')^2}{2l^2}}\right),
\end{equation}
where $\sigma = 50$ and $l = 0.10$. 750 samples are used for training, while 5000 samples are used for testing. One hundred samples are used to calibrate the uncertainty bounds. The system is simulated for two seconds and displacements are recorded at a sampling frequency of 50 Hz. The branch and trunk network consists of four layers each, with 75 nodes in every layer. ReLU activation is used after the first and second layers of both networks, and in the neuroscience-inspired spiking variants of the networks, the ReLU activations in the branch network are replaced by VSN layers. To train the CoNBONet, a learning rate of 0.001 is selected, Adam is used as the optimizer and the network is trained for 20000 iterations with parameters retained for the best iteration in terms of performance gauged using loss function. 

\begin{figure}[ht!]
    \centering
    \includegraphics[width=0.75\linewidth]{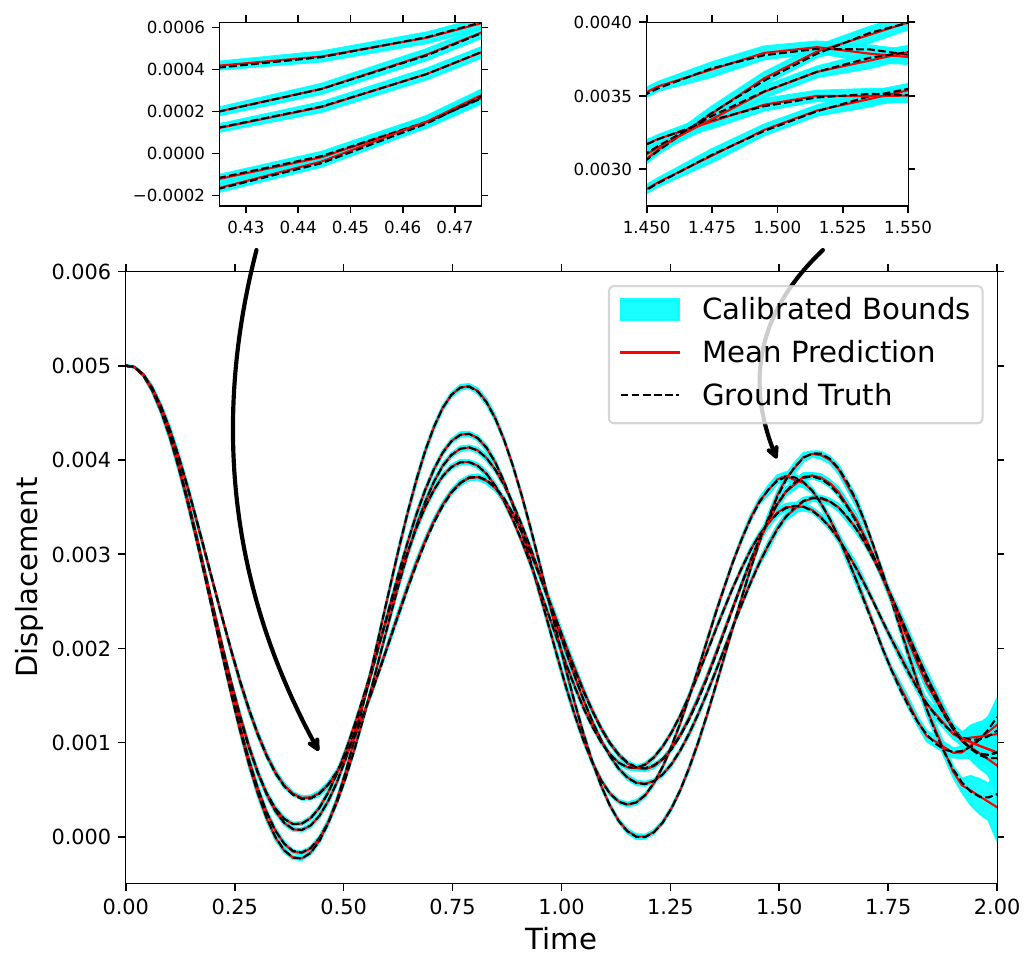}
    \caption{Mean prediction from the CoNBONet compared against the ground truth with calibrated confidence intervals shown as the blue patch, for example E-I.}
    \label{fig: example 1 pred ground truth comparison}
\end{figure}
Fig. \ref{fig: example 1 pred ground truth comparison} shows the mean predictions of CoNBONet compared against the ground truth. As can be seen, the mean prediction closely follows the ground truth and the calibrated uncertainty bounds give good coverage for the ground truth. Normalized Mean Square Error (NMSE) is used as the metric to gauze the performance of networks. An NMSE of $ 1\times10^ {- 4} $ is observed when using CoNBONet and BONet, demonstrating no loss in accuracy when employing VSN layers to introduce sparsity in communication. 
\begin{table}[ht!]
\caption{Spiking activity (\%) for the two layers (A1 and A2) of branch network when using CoNBONet for example E-I.}
\label{tab: spk_values Example 1}
\centering
\begin{tabular}{ccc}
\toprule
Layer & {A1} & {A2} \\ \midrule
1\textsuperscript{st} DOF & 14.70 & 18.20 \\\bottomrule
\end{tabular}
\end{table}
Table \ref{tab: spk_values Example 1} shows the spiking activity in various layers of the CoNBONet. As can be seen, in the current example, we see similar NMSE with and without VSN layers, however, when using VSN layers in branch net, the spiking activity observed is well below 100\%.
\begin{table}[ht!]
\centering
\caption{Coverage provided by the confidence intervals generated using various network in example E-I.}
\label{tab: coverage example E-I}
\begin{tabular}{lccccc}
\toprule
Model & \multicolumn{3}{c}{Coverage (\%)} & \multicolumn{2}{c}{Number of time steps with coverage} \\
 & Avg. & Min. & Max. & $<95\%$ & $\geq95\%$ \\
\midrule
BONet & 93.98 & 81.46 & 100.00 & 53 & 47 \\
CoBONet & 99.73 & 98.68 & 100.00 & 0 & 100 \\
NBONet & 97.56 & 89.00 & 99.98 & 2 & 98 \\
CoNBONet & 97.76 & 96.92 & 99.86 & 0 & 100 \\
\bottomrule
\end{tabular}
\end{table}
Table \ref{tab: coverage example E-I} shows the coverage provided by the uncertainty bounds. As can be seen, the calibrated bounds obtained using CoNBONet provide $\geq 95\%$ coverage at all time steps of the predicted response with only 100 samples used while calibration. 

\begin{figure}
    \centering
    \includegraphics[width=\linewidth]{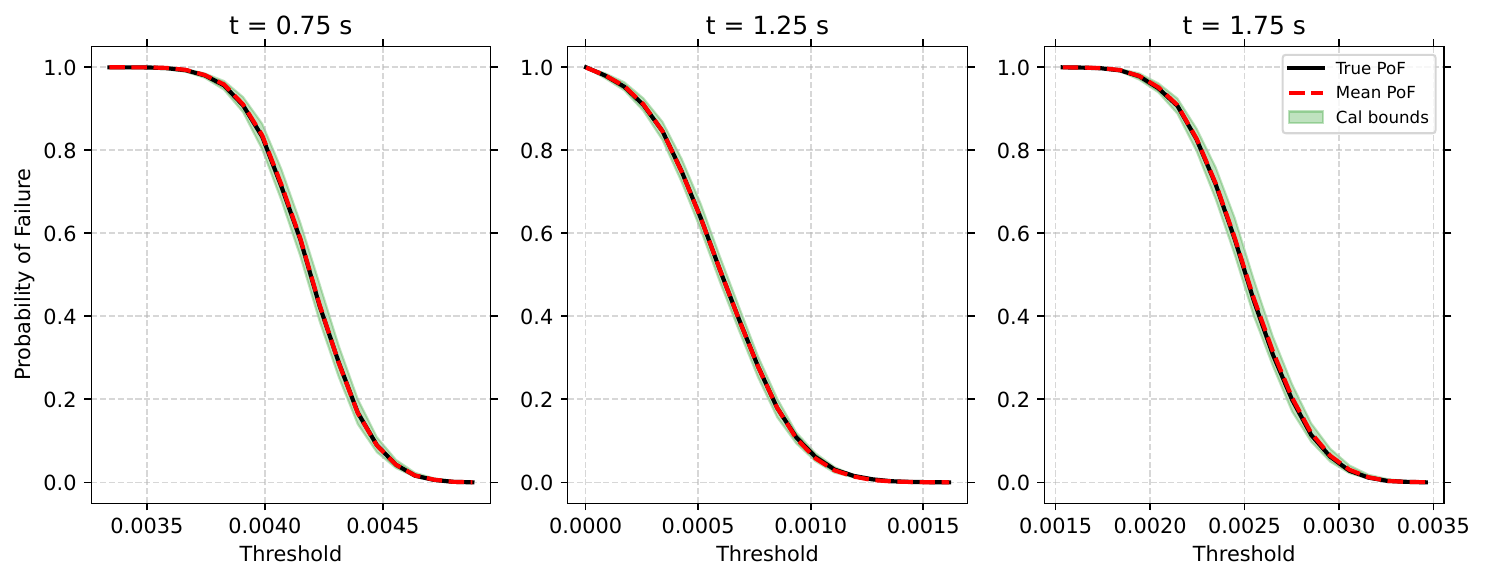}
    \caption{Probability of failure at different time steps in example E-I.}
    \label{fig: POF E-I}
\end{figure}
Fig. \ref{fig: POF E-I} shows the Probability of Failure (PoF) obtained at different time steps corresponding to different threshold values at each time step. As can be seen, the PoF obtained using CoNBONet predictions closely follows the True PoF obtained using MCS results. The results shown correspond to different threshold values; however, in a practical scenario, the system is checked against a single threshold value at a particular time step. 
\subsection{Example II: Multi DOF system with nonlinear oscillator}
In this example, a five DOF system with Duffing oscillator is considered. The Duffing oscillator has been extensively utilized in nonlinear dynamics research to represent systems with cubic stiffness nonlinearity, and it serves as a canonical model for analyzing the complex dynamic behavior of structures, resonators, and vibration isolation devices. The governing equation of motion for the system is defined as, 
\begin{equation}
\begin{matrix}
      m_1\ddot x_1+c_1\dot x_1+c_2(\dot x_1-\dot x_2)+k_1x_1+k_2(x_1-x_2)+\alpha_{do}x_1^3= -m_1f\\
      
      m_2\ddot x_2+c_2(\dot x_2-\dot x_1)+c_3(\dot x_2-\dot x_3)+k_2( x_2-x_1)+k_3( x_2-x_3) = -m_2f\\
      
      m_3\ddot x_3+c_3(\dot x_3-\dot x_2)+c_4(\dot x_3-\dot x_4)+k_3( x_3-x_2)+k_4( x_3-x_4) = -m_3f\\
      
      m_4\ddot x_4+c_4(\dot x_4-\dot x_3)+c_5(\dot x_4-\dot x_5)+k_4( x_4-x_3)+k_5( x_4-x_5) = -m_4f\\
      
      m_5\ddot x_5+c_5(\dot x_5-\dot x_4)+k_5( x_5-x_4) = -m_5f,  
    \end{matrix}
\end{equation}
\begin{table}[ht!]
\centering
\caption{System parameters for Case-II: five-DOF (5-DOF) Duffing oscillator.}
\label{tab:5dof_system_parameters}
\begin{tabular}{ccc}
\toprule
{Mass (kg)} & {Stiffness (N/m)} & {Damping (Ns/m)} \\ 
\midrule
$m_1 = 10$   & $k_1 = 10000$ & $c_1 = 100$ \\ 
$m_2 = 10$   & $k_2 = 10000$ & $c_2 = 100$  \\ 
$m_3 = 9$    & $k_3 = 9000$  & $c_3 = 90$   \\ 
$m_4 = 9$    & $k_4 = 9000$  & $c_4 = 90$   \\ 
$m_5 = 7.5$  & $k_5 = 7500$  & $c_5 = 75$   \\ 
\bottomrule
\end{tabular}
\end{table}
where $m_i$, $c_i$ and $k_i$ are the mass, damping coefficient and stiffness at $i$\textsuperscript{th} DOF, values for which are given in Table \ref{tab:5dof_system_parameters}. $\alpha=100$ here is the nonlinear stiffness coefficient of the Duffing oscillator. Initial conditions for data simulation are taken as $x_0 = 0.01m$, and $\dot x_0 = 0.05m/s$. Same initial conditions are considered at all DOFs. 
The external excitation force applied to the system can be expressed as
\begin{equation}
f(t) = \sum_{i=1}^{n_s} a_{s_i} \sin(f_{s_i} t) + \sum_{i=1}^{n_c} a_{c_i} \cos(f_{c_i} t),
\label{equation:force_fourier}
\end{equation}
where $a_{s_i}$ and $a_{c_i}$ represent the amplitudes of the sinusoidal components, and $f_{s_i}$ and $f_{c_i}$ denote their corresponding frequencies.
The parameters $a_s$, $a_c$, $f_s$, and $f_c$ are treated as uniformly distributed random variables to model the stochastic characteristics of the applied excitation.
The integers $n_s$ and $n_c$ correspond to the numbers of sine and cosine terms, respectively, with the total number of Fourier terms given by $n = n_s + n_c$. In this example, $n_s = n_c = 10$.

500 samples are used for training while 5000 samples are used for testing. 135 samples are used for calibrating the uncertainty bounds. The system is simulated for two seconds and displacements are recorded at a sampling frequency of 50 Hz. The branch and trunk network consists of four layers each with 100 nodes in each layer. Similar to previous example, ReLU activation is used after first and second layer of both networks and in the spiking variants of the networks, the ReLU activations in the branch network are replaced by VSN layers. To train the CoNBONet, a learning rate of 0.0001 is selected, Adam is used as the optimizer and the network is trained for 35000 iterations with parameters retained for the best iteration in terms of performance gauged using loss function. It should be noted that system response corresponding to each DOF will be learned separately using its own CoNBONet model. 

\begin{figure}[ht!]
    \centering

    \begin{subfigure}[t]{0.45\textwidth}
        \centering
        \includegraphics[width=\textwidth]{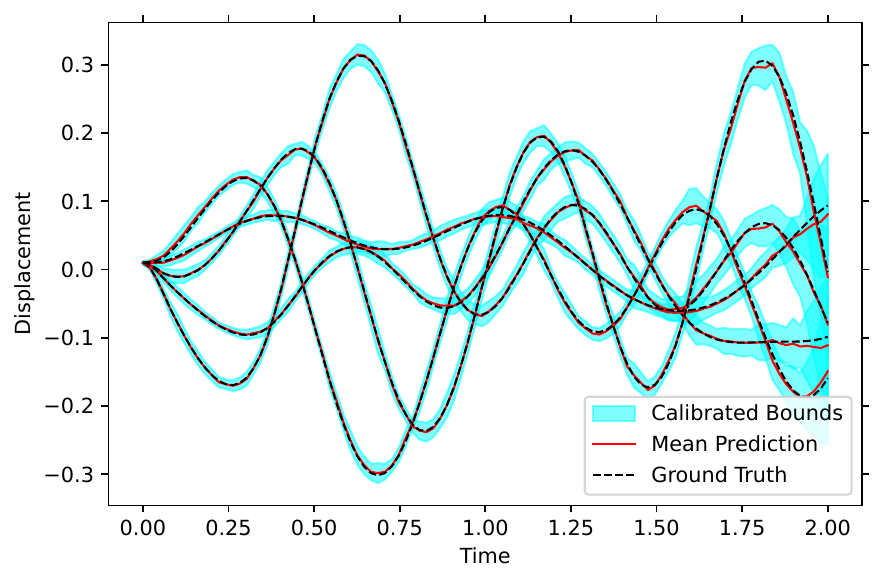}
        \caption{1\textsuperscript{st} DOF}
    \end{subfigure}
    \hfill
    \begin{subfigure}[t]{0.45\textwidth}
        \centering
        \includegraphics[width=\textwidth]{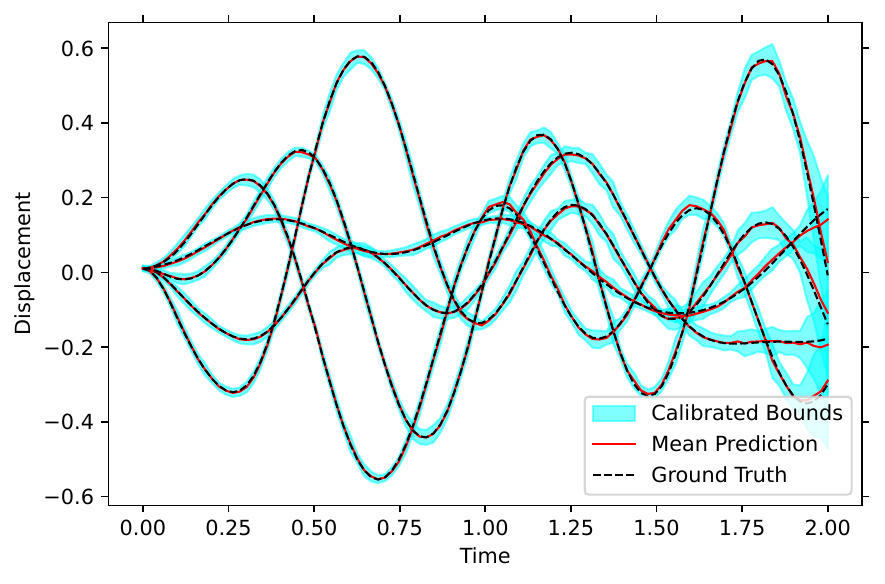}
        \caption{2\textsuperscript{nd} DOF}
    \end{subfigure}

    \vspace{1em} 

    \begin{subfigure}[t]{0.45\textwidth}
        \centering
        \includegraphics[width=\textwidth]{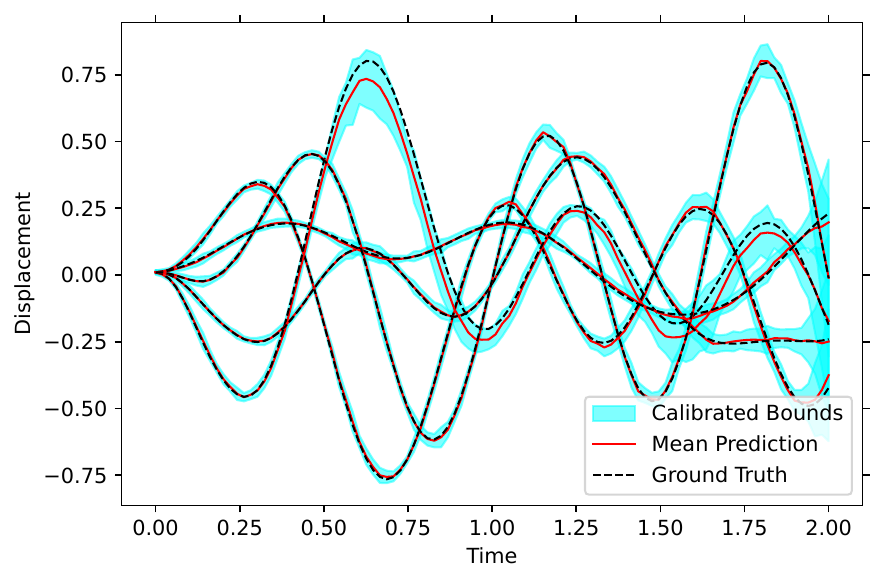}
        \caption{3\textsuperscript{rd} DOF}
    \end{subfigure}
    \hfill
    \begin{subfigure}[t]{0.45\textwidth}
        \centering
        \includegraphics[width=\textwidth]{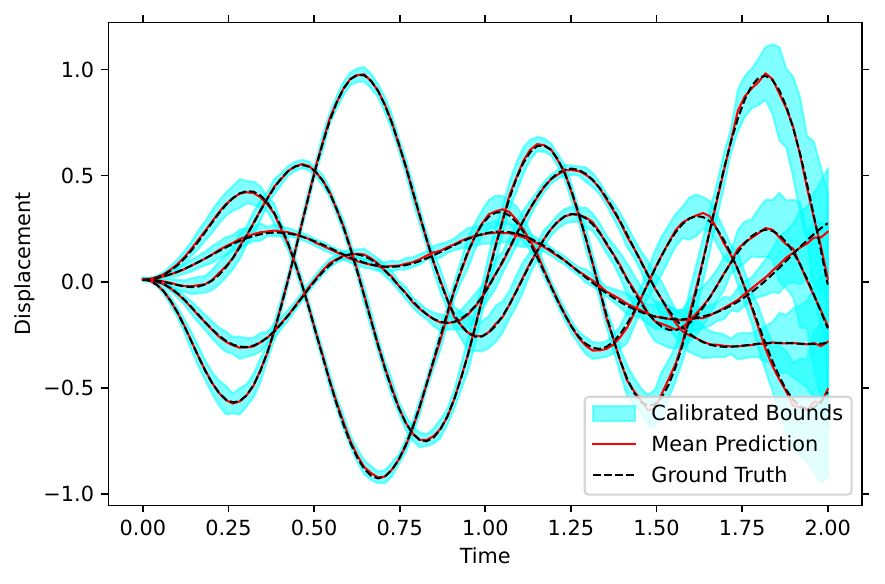}
        \caption{4\textsuperscript{th} DOF}
    \end{subfigure}
    \hfill
    \begin{subfigure}[t]{0.45\textwidth}
        \centering
        \includegraphics[width=\textwidth]{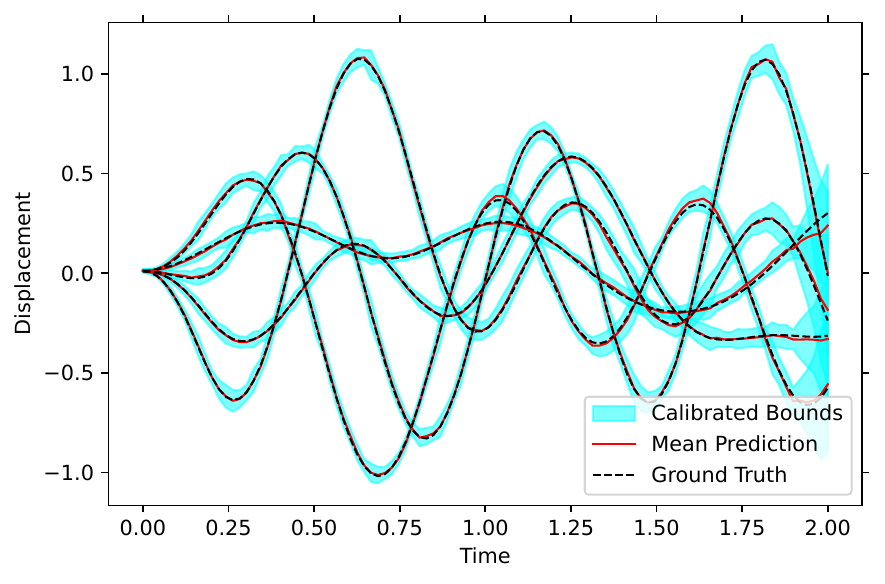}
        \caption{5\textsuperscript{th} DOF}
    \end{subfigure}

    \caption{Mean prediction from the CoNBONet compared against the ground truth with calibrated confidence intervals shown as the blue patch, for example E-II.}
    \label{fig: E-II pre vs gt and bounds}
\end{figure}
\begin{table}[ht!]
\centering
\caption{NMSE for each degree of freedom obtained when comparing network predictions against the ground truth in E-II.}
\label{tab: E-II NMSE}
\begin{tabular}{l c c c c c}
\toprule
 Network & 1\textsuperscript{st} DOF & 2\textsuperscript{nd} DOF & 3\textsuperscript{rd} DOF & 4\textsuperscript{th} DOF & 5\textsuperscript{th} DOF \\
\midrule
BONet & 0.0018 & 0.0019 & 0.0027 & 0.0009 & 0.0018 \\
CoNBONet & 0.0048 & 0.0060 & 0.0022 & 0.0044 & 0.0051 \\
\bottomrule
\end{tabular}
\end{table}
Fig. \ref{fig: E-II pre vs gt and bounds} presents the CoNBONet predictions in comparison with the ground truth. For various DOFs, the predicted responses from their respective CoNBONet model closely match the ground truth. In regions where the mean predictions deviate slightly, the associated uncertainty bounds effectively capture and encompass the ground truth.
NMSE values from Table \ref{tab: E-II NMSE} further strengthens the visual evidence as observed in Fig. \ref{fig: E-II pre vs gt and bounds}.
\begin{table}[ht!]
\centering
\caption{Spiking activity (\%) for the two layers (A1 and A2) of the branch network in CoNBONet (Example E-II).}
\label{tab: E-II spk values}

\begin{tabular}{lccccc}
\toprule
{Layer} & 1\textsuperscript{st} DOF & 2\textsuperscript{nd} DOF & 3\textsuperscript{rd} DOF & 4\textsuperscript{th} DOF & 5\textsuperscript{th} DOF \\ 
\midrule
A1 & 11.5 & 12.5 & 14.0 & 15.0 & 16.1 \\ 
A2 & 17.5 & 14.0 & 16.2 & 22.5 & 16.4 \\ 
\bottomrule
\end{tabular}
\end{table}
Spiking activity in Table \ref{tab: E-II spk values} show that CoNBONet is able to produce accurate mean predictions despite spiking activity staying between 10\% and 25\%.
\begin{table}[ht!]
\centering
\caption{Coverage provided by the confidence intervals generated using various network in example E-II.}
\label{tab: E-II coverage}
\begin{tabular}{l l cccc}
\toprule
 & & \multicolumn{3}{c}{Coverage (\%)} & \multicolumn{1}{c}{Number of time steps with coverage} \\
DOF & Model & AvgCov & MinCov & MaxCov & $<95\%$ / $\geq95$ (\%) \\
\midrule
1\textsuperscript{st} & BONet & 95.30 & 83.72 & 99.94 & 29 / 71 \\
 & CoBONet & 99.23 & 98.18 & 99.96 & 0 / 100 \\
 & NBONet & 98.37 & 97.54 & 100.00 & 0 / 100 \\
 & CoNBONet & 98.52 & 97.24 & 99.98 & 0 / 100 \\
\hline
2\textsuperscript{nd} & BONet & 92.97 & 75.78 & 100.00 & 59 / 41 \\
 & CoBONet & 98.71 & 97.26 & 99.92 & 0 / 100 \\
 & NBONet & 98.90 & 97.26 & 100.00 & 0 / 100 \\
 & CoNBONet & 98.87 & 98.32 & 99.98 & 0 / 100 \\
\hline
3\textsuperscript{rd} & BONet & 95.71 & 80.20 & 100.00 & 30 / 70 \\
 & CoBONet & 99.06 & 98.10 & 99.80 & 0 / 100 \\
 & NBONet & 99.35 & 97.80 & 100.00 & 0 / 100 \\
 & CoNBONet & 99.34 & 98.52 & 99.96 & 0 / 100 \\
\hline
4\textsuperscript{th} & BONet & 94.83 & 78.30 & 100.00 & 34 / 66 \\
 & CoBONet & 99.66 & 98.14 & 100.00 & 0 / 100 \\
 & NBONet & 98.26 & 96.92 & 100.00 & 0 / 100 \\
 & CoNBONet & 98.84 & 97.94 & 99.96 & 0 / 100 \\
\hline
5\textsuperscript{th} & BONet & 91.90 & 74.06 & 100.00 & 89 / 11 \\
 & CoBONet & 97.40 & 95.22 & 99.96 & 0 / 100 \\
 & NBONet & 97.70 & 96.06 & 100.00 & 0 / 100 \\
 & CoNBONet & 98.18 & 96.74 & 100.00 & 0 / 100 \\
\bottomrule
\end{tabular}
\end{table}
Table \ref{tab: E-II coverage} shows the coverage provided by the uncertainty bounds obtained using various networks. In this example, NBONet provides near-nominal empirical coverage; however, its conformalized version, the CoNBONet, performs as expected and provides required coverage at all time points. 

\begin{figure}[ht!]
    \centering

    \begin{subfigure}[t]{0.45\textwidth}
        \centering
        \includegraphics[width=\textwidth]{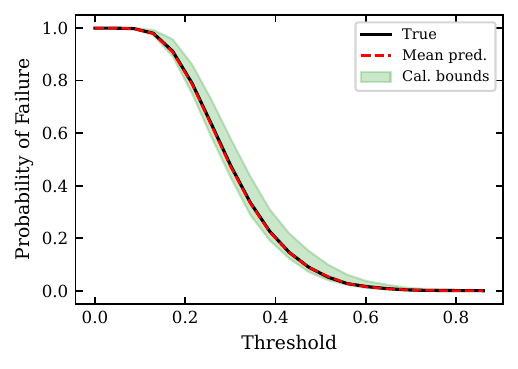}
        \caption{1\textsuperscript{st} DOF}
    \end{subfigure}
    \hfill
    \begin{subfigure}[t]{0.45\textwidth}
        \centering
        \includegraphics[width=\textwidth]{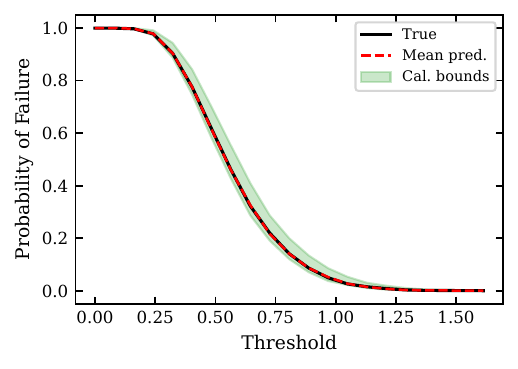}
        \caption{2\textsuperscript{nd} DOF}
    \end{subfigure}

    \vspace{1em} 

    \begin{subfigure}[t]{0.45\textwidth}
        \centering
        \includegraphics[width=\textwidth]{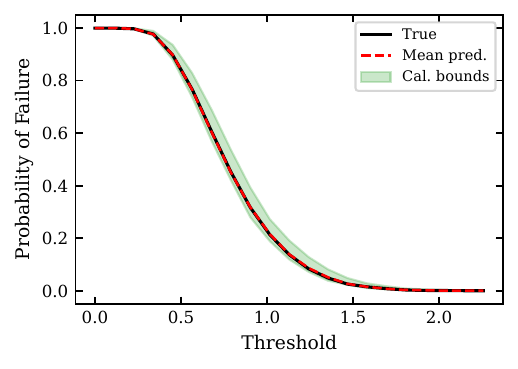}
        \caption{3\textsuperscript{rd} DOF}
    \end{subfigure}
    \hfill
    \begin{subfigure}[t]{0.45\textwidth}
        \centering
        \includegraphics[width=\textwidth]{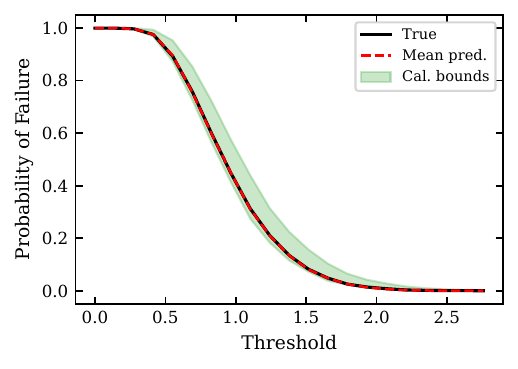}
        \caption{4\textsuperscript{th} DOF}
    \end{subfigure}
    \hfill
    \begin{subfigure}[t]{0.45\textwidth}
        \centering
        \includegraphics[width=\textwidth]{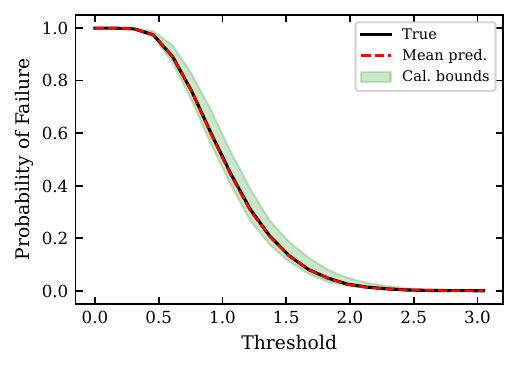}
        \caption{5\textsuperscript{} DOF}
    \end{subfigure}

    \caption{PoF obtained using FTTF analysis at various DOFs in E-II. The PoF is computed corresponding to different threshold vales at each DOF.}
    \label{fig: E-II PoF five_figures}
\end{figure}
Fig. \ref{fig: E-II PoF five_figures} shows the PoF obtained using FTTF analysis at various DOFs. The PoF is computed corresponding to different threshold vales at each DOF. As expected, since the mean predictions of the CoNBONet architecture closely followed the ground truth the PoF obtained using the same also follow similar trend. Furthermore, the advantage of using CoNBONet architecture is that along with mean prediction of PoF, we obtain a uncertainty bound that provides user auxiliary information that can prove to be instrumental in decision making tasks.
%
\begin{table}[ht!]
\centering
\caption{PoF obtained using FTTF analysis at various DOFs in E-II. The PoF is computed at each DOF, correspondin to a single set of threshold values.}
\label{tab: pof_per_dof}
\begin{tabular}{lccccc}
\toprule
 & DOF 0 & DOF 1 & DOF 2 & DOF 3 & DOF 4 \\
\midrule
Threshold & 0.25 & 0.48 & 0.67 & 0.82 & 0.91\\
True PoF & 0.633 & 0.617 & 0.607 & 0.599 & 0.596 \\
$\underset{\text{Bounds}}{\text{Mean PoF}}$ & 
$\underset{[0.584,\,0.736]}{0.628}$ &
$\underset{[0.581,\,0.694]}{0.614}$ &
$\underset{[0.569,\,0.685]}{0.603}$ &
$\underset{[0.563,\,0.718]}{0.596}$ &
$\underset{[0.557,\,0.694]}{0.593}$ \\
\bottomrule
\end{tabular}
\end{table}
Table \ref{tab: pof_per_dof} shows the PoF corresponding to a single set of threshold values. The results produced further strengthen the results shown in Fig. \ref{fig: E-II PoF five_figures}.
\subsection{Example III: 76 DOF engineering system}
To further evaluate the robustness and scalability of the proposed DeepONet surrogate framework, it is applied to the well-known ASCE 76-story benchmark problem introduced by Yang et. al. \cite{yang2004benchmark}. This benchmark represents a realistic high rise office building, providing a standardized platform for testing. The linear system parameters, mass, stiffness, and damping matrices, are adopted from the original 76-DOF benchmark model \cite{yang2004benchmark,nayek2019gaussian}, which was designed as a \(306.1\,\text{m}\) tall reinforced concrete tower located in Melbourne, Australia. The building consists of a \(42\,\text{m} \times 42\,\text{m}\) outer concrete frame surrounding a \(21\,\text{m} \times 21\,\text{m}\) central core. The exterior frame is supported by square columns spaced at \(6.5\,\text{m}\) intervals and connected through spandrel beams at each floor, ensuring structural continuity.

For the present study, the linear 76-DOF system is modified by introducing a Bouc-Wen hysteretic oscillator at the first DOF to capture nonlinear restoring force behavior. This modification allows investigation of the surrogate model's ability to represent nonlinear dynamic responses within a high-dimensional structural system. The governing equations of motion are accordingly updated to include the Bouc-Wen nonlinearity. The external excitation force applied to the system is in accordance with Eq. \eqref{equation:force_fourier}. The CoNBONet architecture adopted is similar to the previous example, with nodes in each layer set to 86. 40000 iterations are used to train the model and 350 training samples are used for the same. 

\begin{figure}[ht!]
    \centering

    \begin{subfigure}[t]{0.45\textwidth}
        \centering
        \includegraphics[width=\textwidth]{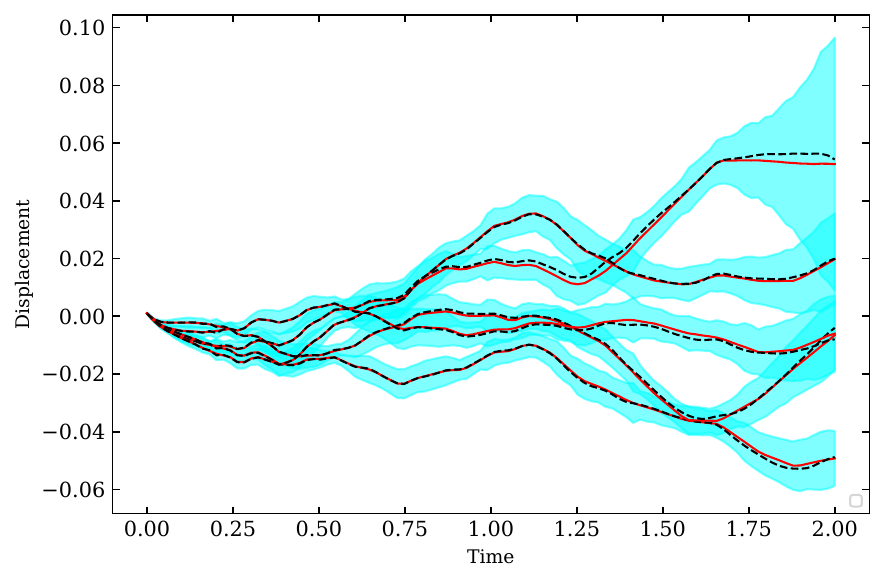}
        \caption{5\textsuperscript{th} DOF}
    \end{subfigure}
    \hfill
    \begin{subfigure}[t]{0.45\textwidth}
        \centering
        \includegraphics[width=\textwidth]{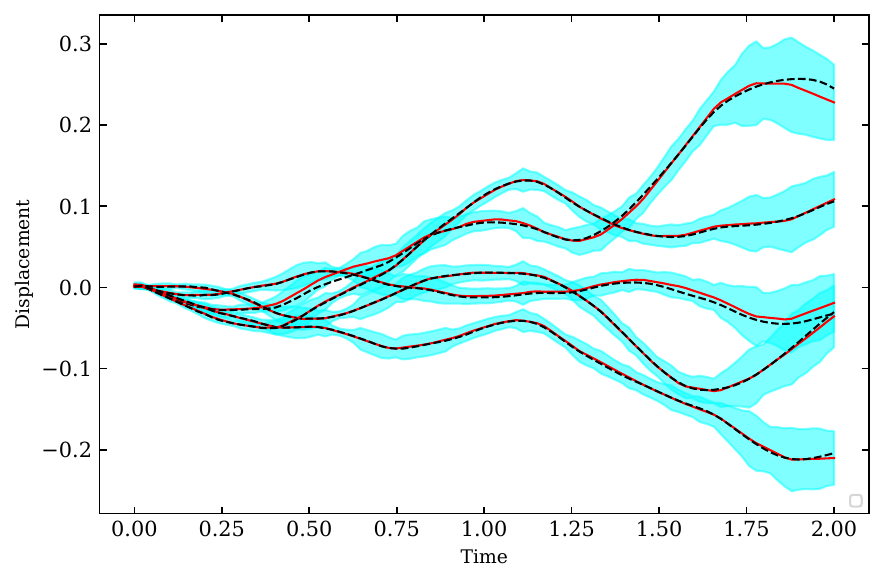}
        \caption{15\textsuperscript{th} DOF}
    \end{subfigure}

    \vspace{1em} 

    \begin{subfigure}[t]{0.45\textwidth}
        \centering
        \includegraphics[width=\textwidth]{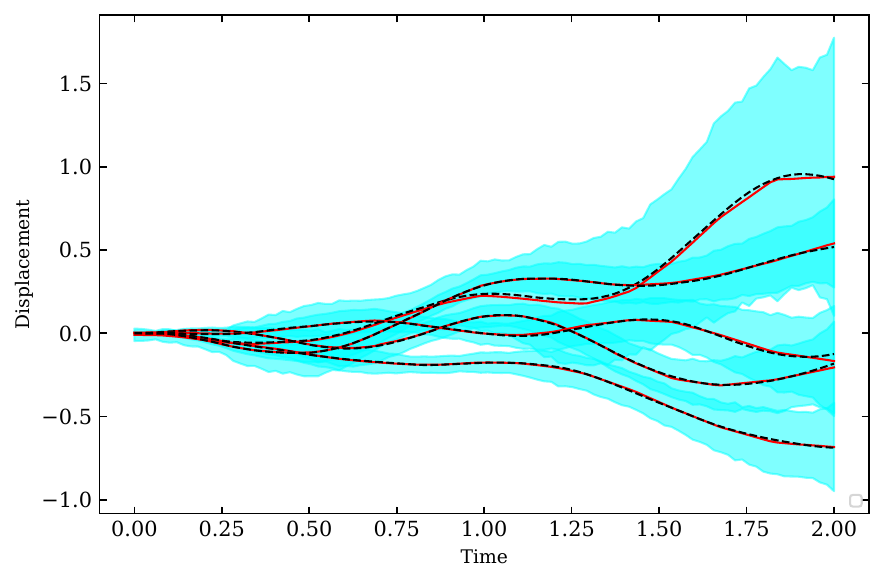}
        \caption{35\textsuperscript{th} DOF}
    \end{subfigure}
    \hfill
    \begin{subfigure}[t]{0.45\textwidth}
        \centering
        \includegraphics[width=\textwidth]{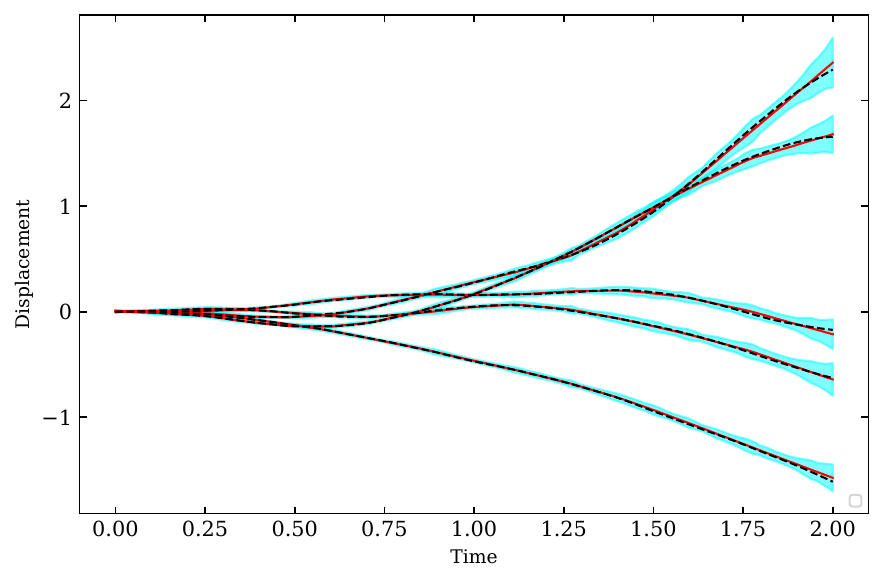}
        \caption{65\textsuperscript{th} DOF}
    \end{subfigure}
    \hfill
    \begin{subfigure}[t]{0.45\textwidth}
        \centering
        \includegraphics[width=\textwidth]{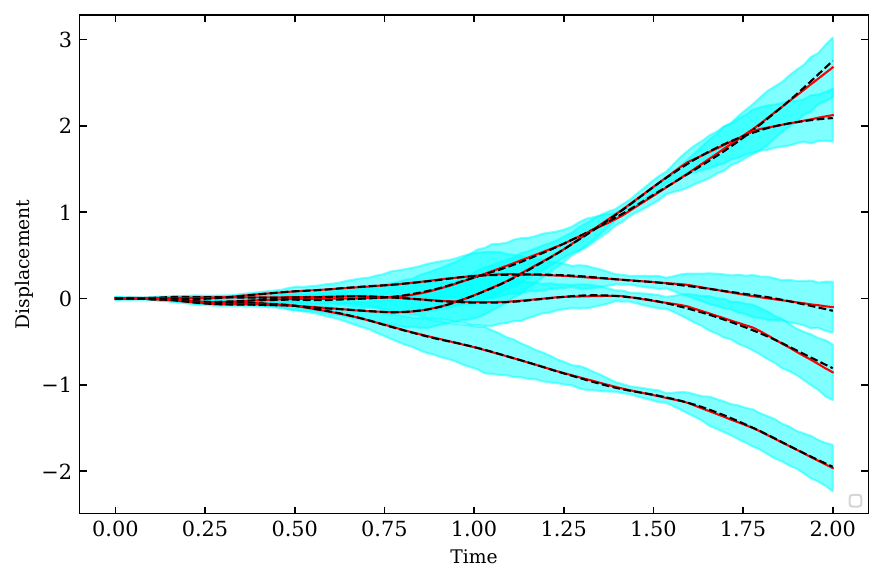}
        \caption{75\textsuperscript{th} DOF}
    \end{subfigure}

    \caption{Mean prediction from the CoNBONet compared against the ground truth with calibrated confidence intervals shown as the blue patch, for example E-III.}
    \label{fig: E-III pred vs ground truth}
\end{figure}

\begin{table}[ht!]
\caption{NMSE for each degree of freedom obtained when comparing network predictions against the ground truth in E-III.}
\label{tab: E-III NMSE}
\centering
\begin{tabular}{l c c c c c}
\toprule
Network & 5\textsuperscript{th} DOF & 15\textsuperscript{th} DOF & 35\textsuperscript{th} DOF & 65\textsuperscript{th} DOF & 75\textsuperscript{th} DOF \\
\midrule
BONet & 0.0083 & 0.0027 & 0.0021 & 0.0011 & 0.0004 \\
CoNBONet& 0.0088 & 0.0072 & 0.0125 & 0.1101 & 0.0132 \\
\bottomrule
\end{tabular}
\end{table}

\begin{table}[ht!]
\centering
\caption{Spiking activity (\%) for the two layers (A1 and A2) of the branch network in CoNBONet (Example E-III).}
\label{tab: E-III spk values}
\begin{tabular}{lccccc}
\toprule
{Layer} & 5\textsuperscript{th} DOF & 15\textsuperscript{th} DOF & 35\textsuperscript{th} DOF & 65\textsuperscript{th} DOF & 75\textsuperscript{th} DOF \\ 
\midrule
A1 & 12.4 & 23.0 & 09.5 & 08.2 & 11.4 \\ 
A2 & 31.3 & 15.0 & 24.3 & 09.4 & 08.0 \\ 
\bottomrule
\end{tabular}
\end{table}
Fig. \ref{fig: E-III pred vs ground truth} shows the mean predictions obtained from CoNBONet, compared against the ground truth. Tables \ref{tab: E-III NMSE} and \ref{tab: E-III spk values} shows the NMSE values obtained and the spiking activity in various VSN layers of CoNBONet architecture. The results produced follow a similar trend to the previous examples, wherein the CoNBONet predictions give a good approximation of the ground truth despite sufficiently sparse communication.
%
%
\begin{table}[ht!]
\centering
\caption{Coverage provided by the confidence intervals generated using various network in example E-III.}
\label{tab: E-III coverage}
\begin{tabular}{l l cccc}
\toprule
DOF & Model & AvgCov & MinCov & MaxCov & $<95$ / $\geq95$ (\%) \\
\midrule
5\textsuperscript{th} & BONet & 85.91 & 76.36 & 100.00 & 86 / 14.00 \\
 & CoBONet & 98.66 & 97.54 & 100.00 & 0 / 100.00 \\
 & NBONet & 82.72 & 54.00 & 99.80 & 88 / 12.00 \\
 & CoNBONet & 99.02 & 97.62 & 99.72 & 0 / 100.00 \\
\midrule
15\textsuperscript{th} & BONet & 89.97 & 81.18 & 99.96 & 73 / 27.00 \\
 & CoBONet & 98.95 & 98.12 & 99.96 & 0 / 100.00 \\
 & NBONet & 90.21 & 79.28 & 99.54 & 81 / 19.00 \\
 & CoNBONet & 97.73 & 95.66 & 99.80 & 0 / 100.00 \\
\midrule
35\textsuperscript{th} & BONet & 91.94 & 77.72 & 100.00 & 65 / 35.00 \\
 & CoBONet & 98.26 & 96.78 & 99.98 & 0 / 100.00 \\
 & NBONet & 86.95 & 72.34 & 99.90 & 83 / 17.00 \\
 & CoNBONet & 98.65 & 97.44 & 99.74 & 0 / 100.00 \\
\midrule
65\textsuperscript{th} & BONet & 93.92 & 84.24 & 100.00 & 54 / 46.00 \\
 & CoBONet & 99.38 & 97.60 & 100.00 & 0 / 100.00 \\
 & NBONet & 96.18 & 83.90 & 99.94 & 23 / 77.00 \\
 & CoNBONet & 96.88 & 95.04 & 99.82 & 0 / 100.00 \\
\midrule
75\textsuperscript{th} & BONet & 95.75 & 85.56 & 100.00 & 37 / 63.00 \\
 & CoBONet & 98.90 & 97.14 & 100.00 & 0 / 100.00 \\
 & NBONet & 95.71 & 84.64 & 100.00 & 40 / 60.00 \\
 & CoNBONet & 98.40 & 97.48 & 99.96 & 0 / 100.00 \\
\bottomrule
\end{tabular}
\end{table}
Table \ref{tab: E-III coverage} shows the coverage provided compared against the other networks. As can be seen, the conformalized versions of BONet and NBONet irrevocably outperform their vanilla counterparts and the additional advantage of sparse communication is observed when using CoNBONet.

Reliability analysis in the current example is carried out at the last DOF i.e. $76$\textsuperscript{th} DOF.
\begin{figure}[ht!]
    \centering
    \includegraphics[width=0.6\linewidth]{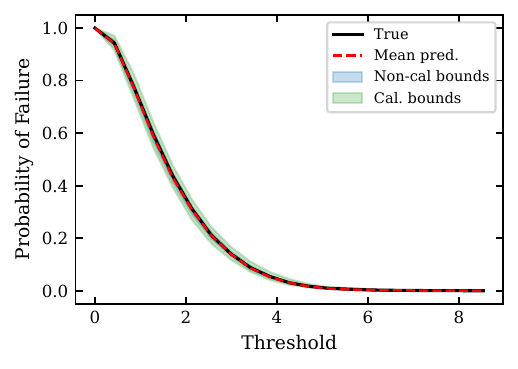}
    \caption{PoF obtained using FTTF analysis at 76\textsuperscript{th} DOF in E-III. The PoF is computed corresponding to different threshold vales.}
    \label{fig: E-III 76 PoF}
\end{figure}
Fig. \ref{fig: E-III 76 PoF} shows the PoF obtained using FTTF analysis at 76\textsuperscript{th} DOF. The PoF is computed corresponding to different threshold vales. As can be seen, the PoF obtained using mean predictions of CoNBONet closely follow the ground truth with additional information available in the form of uncertainty bounds.
\section{Conclusion}\label{section: conclusion}
Time-dependent reliability analysis of nonlinear dynamical systems under stochastic excitations poses significant computational challenges, primarily due to the need for repeated numerical integrations across numerous random realizations. Conventional methods such as Monte Carlo simulation, while accurate, are computationally prohibitive for high-fidelity models and real-time applications. 
To overcome these limitations, this study introduced \textbf{CoNBONet}, a {\textbf{Conformalized} \textbf{Neuroscience}\textbf{-}\textbf{inspired} \textbf{Bayesian} \textbf{Operator} \textbf{Network}}, as a fast, scalable, energy-efficient, and uncertainty-aware surrogate modeling framework for reliability analysis. CoNBONet integrates the expressive power of operator learning through a Bayesian DeepONet backbone with the \textit{energy efficiency of Variable Spiking Neurons (VSNs)} and the \textit{theoretical calibration guarantees of split conformal prediction}. This combination enables efficient operator learning with sparse, event-driven computation and reliable uncertainty quantification.

The key observations and highlights from the numerical investigations are summarized as follows:
\begin{itemize}
    \item \textbf{Preserved Predictive Accuracy:} Across all examples, from a single DOF Bouc-Wen system to a 76-DOF nonlinear building system, CoNBONet achieved predictive fidelity comparable to the gold-standard Monte Carlo simulations, demonstrating its effectiveness in capturing nonlinear dynamics under stochastic loading.  
    \item \textbf{Energy-Efficient Computation:} The incorporation of VSN layers reduced spiking activity to below 25\% on average, implying substantial savings in MAC operations with marginal effect on accuracy, thereby promoting suitability for embedded or neuromorphic deployment.  
    \item \textbf{Reliable Uncertainty Quantification:} Through conformal calibration, CoNBONet consistently achieved the desired coverage level ($\geq$95\%) across all time steps and degrees of freedom, ensuring trustworthy probabilistic predictions critical for safety assessment.  
    \item \textbf{Accurate Reliability Estimation:} The time-dependent probability of failure (PoF) computed using CoNBONet closely matched the reference Monte Carlo results across all test cases. The calibrated uncertainty bounds around the PoF estimates provided meaningful confidence intervals, reinforcing the model's capability for robust reliability assessment under uncertainty. 
\end{itemize}
Overall, CoNBONet offers an efficient and energy-aware surrogate modeling framework for time-dependent reliability analysis, marking a significant step forward in striking a balance between computational efficiency and predictive credibility.  

Despite these promising outcomes, certain limitations remain. The current framework relies on supervised training data, which can be expensive to generate for extremely high-dimensional stochastic systems. Moreover, while VSNs theoretically enhance energy efficiency, their event-driven dynamics are yet to be tested on real-life neuromorphic hardware. Future work can focus on integrating physics-informed priors and deploying CoNBONet on neuromorphic hardware to further enhance its efficiency and applicability in large-scale engineering reliability analysis.
\section*{Acknowledgment}
SG acknowledges the financial support received from the Ministry of Education, India, in the form of the Prime Minister's Research Fellows (PMRF) scholarship. SC acknowledges the financial support received from the Anusandhan National Research Foundation (ANRF) via grant no. CRG/2023/007667 and from the Ministry of Port and Shipping via letter no. ST-14011/74/MT (356529).

\end{document}